\documentclass[runningheads]{llncs}
\usepackage{graphicx}

\usepackage{tikz}
\usepackage{comment}
\usepackage{amsmath,amssymb} 
\usepackage{color}

\usepackage{enumitem}
\setitemize{noitemsep,topsep=0pt,parsep=0pt,partopsep=0pt}

\usepackage[accsupp]{axessibility}  

\makeatletter
\@namedef{ver@everyshi.sty}{}
\makeatother
\usetikzlibrary{spy}
\usepackage{cleveref}
\usepackage{pgfplots} 
\usetikzlibrary{patterns}
\usepackage{pifont}
\newcommand{\xmark}{\ding{55}}

\begin{document}
\pagestyle{headings}
\mainmatter
\def\ECCVSubNumber{305}  

\title{Event-based Image Deblurring with Dynamic Motion Awareness} 

\titlerunning{Event-based Image Deblurring with Dynamic Motion Awareness}
%
\author{Patricia Vitoria \orcidID{0000-0002-2437-0191} \and
Stamatios Georgoulis  \and
Stepan Tulyakov  \and 
Alfredo Bochicchio \and
Julius Erbach \and 
Yuanyou Li}

\authorrunning{P. Vitoria et al.}
 \vspace{-3ex}
\institute{Huawei Technologies, Zurich Research Center \\
\email{patricia.vitoria.carrera@huawei.com, name.surname@huawei.com}}
\maketitle 

\begin{abstract}
Non-uniform image deblurring is a challenging task due to the lack of temporal and textural information in the blurry image itself. 
Complementary information from auxiliary sensors such event sensors are being explored to address these limitations.
The latter can record changes in a logarithmic intensity asynchronously, called events, with high temporal resolution and high dynamic range. 
Current event-based deblurring methods combine the blurry image with events to jointly estimate per-pixel motion and the deblur operator. 
In this paper, we argue that a divide-and-conquer approach is more suitable for this task. 
To this end, we propose to use modulated deformable convolutions, whose kernel offsets and modulation masks are dynamically estimated from events to encode the motion in the scene, while the deblur operator is learned from the combination of blurry image and corresponding events. 
Furthermore, we employ a coarse-to-fine multi-scale reconstruction approach to cope with the inherent sparsity of events in low contrast regions.
Importantly, we introduce the first dataset containing pairs of real RGB blur images and related events during the exposure time.
Our results show better overall robustness when using events, with improvements in PSNR by up to 1.57dB on synthetic data and 1.08 dB on real event data.

\end{abstract}

\section{Introduction}
\label{sec:intro}
Conventional cameras operate by accumulating light over the exposure time and integrating this information to produce an image.
During this time, camera or scene motion may lead to a mix of information across a neighborhood of pixels, resulting in blurry images. 
To account for this effect, some cameras incorporate an optical image stabilizer (OIS), which tries to actively counteract the camera shake.
However, OIS can deal just with small motions and it still fails in the case of scene motion.
To reduce the impact of motion blur, one can alternatively adapt the exposure time according to the motion observed in the scene~\cite{delbracio2021mobile}, albeit at the potential loss of structural information.
In the inevitable case where an image is already blurred, image deblurring methods try to recover the sharp image behind the motion-blurred scene. 
However, restoring a photorealistic sharp image from a single blurry image is an ill-posed problem due to the loss of temporal and textural information.

In the last years, image-based deblurring algorithms have shown impressive advances~\cite{nah2017deep,jin2018learning,tao2018scale,zhang2018dynamic,kupyn2018deblurgan,kupyn2019deblurgan,zhang2019deep,gao2019dynamic,yuan2020efficient,park2020multi,suin2020spatially,huo2021blind,cho2021rethinking,zamir2021multi,chen2021hinet}, but they still struggle while reconstructing high-frequency details.
Recently, together with the intensity camera, a new breed of bio-inspired sensors containing high-frequency details, called event cameras, have been employed in image deblurring.
Event cameras can asynchronously detect local log intensity changes, called \emph{events}, at each pixel independently during the exposure time of a conventional camera with very low latency (1 $\mu s$) and high dynamic range ($\approx$ 140 dB).
Understandably, due to their capacity to capture high temporal resolution and fine-grained motion information, event cameras pose the deblurring problem in a more tractable way.
Despite their great potential, current event-based image deblurring methods come with their own limitations.
\emph{(1)} They are restricted to low-resolution Active Pixel Sensor (APS) intensity frames provided by the event camera~\cite{pan2019bringing,pan2020high,jiang2020learning,haoyu2020learning,sun2021mefnet} or
\emph{(2)} work just with grayscale images or using color events by treating each channel separately, failing to exploit the intra-correlations between image channels~\cite{pan2019bringing,pan2020high,jiang2020learning,haoyu2020learning,sun2021mefnet}.
\emph{(3)} They are either based on physical models~\cite{pan2019bringing,pan2020high,jiang2020learning} that are sensitive to the lack of events in low-contrast regions as well as their noisy nature, or they use the event data just as an additional input to the image without explicitly exploiting the information provided by the events~\cite{haoyu2020learning,zhang2020hybrid}.
\emph{(4)} They do not generalize well to real data~\cite{haoyu2020learning,zhang2020hybrid}.

\begin{figure*}[ht!]
    \centering
    \includegraphics[trim=0 20 0 0,clip,width=0.9\textwidth]{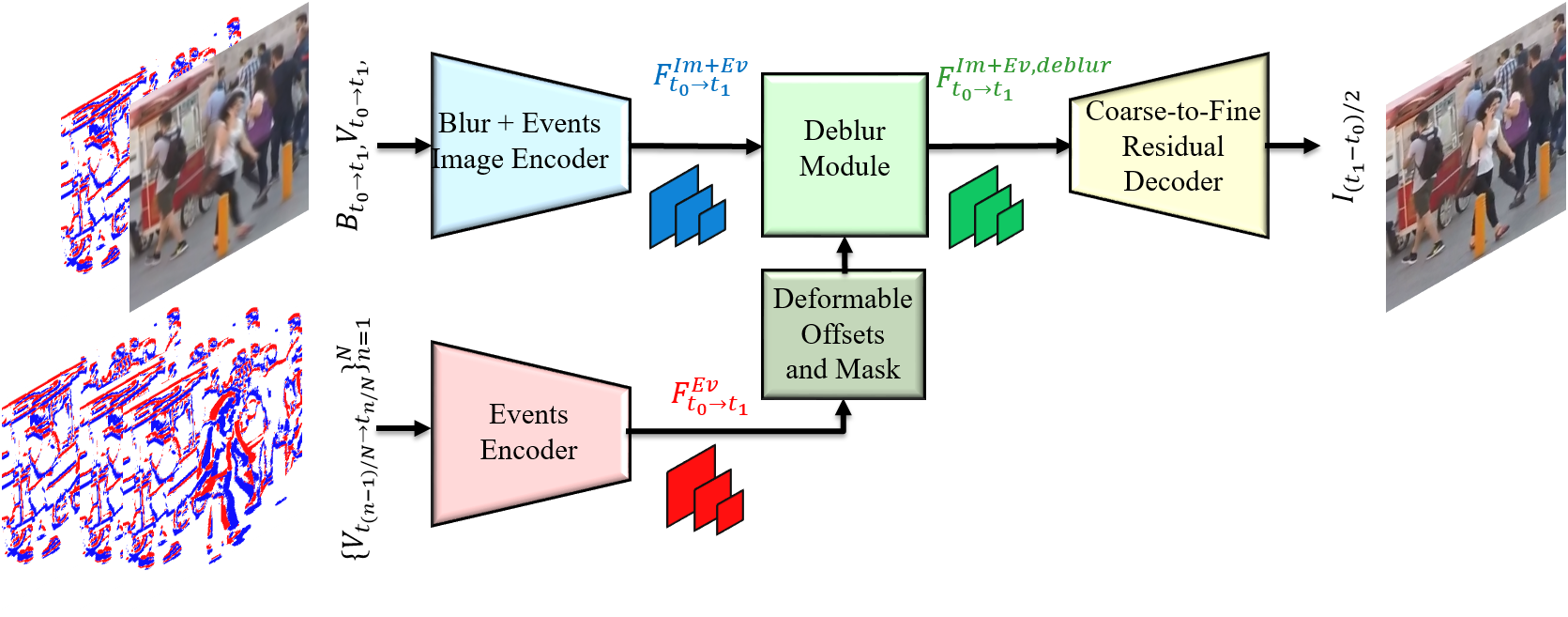} \\
    (a) Proposed Pipeline
    \begin{tabular}{c|c}
        \includegraphics[width=0.45\linewidth]{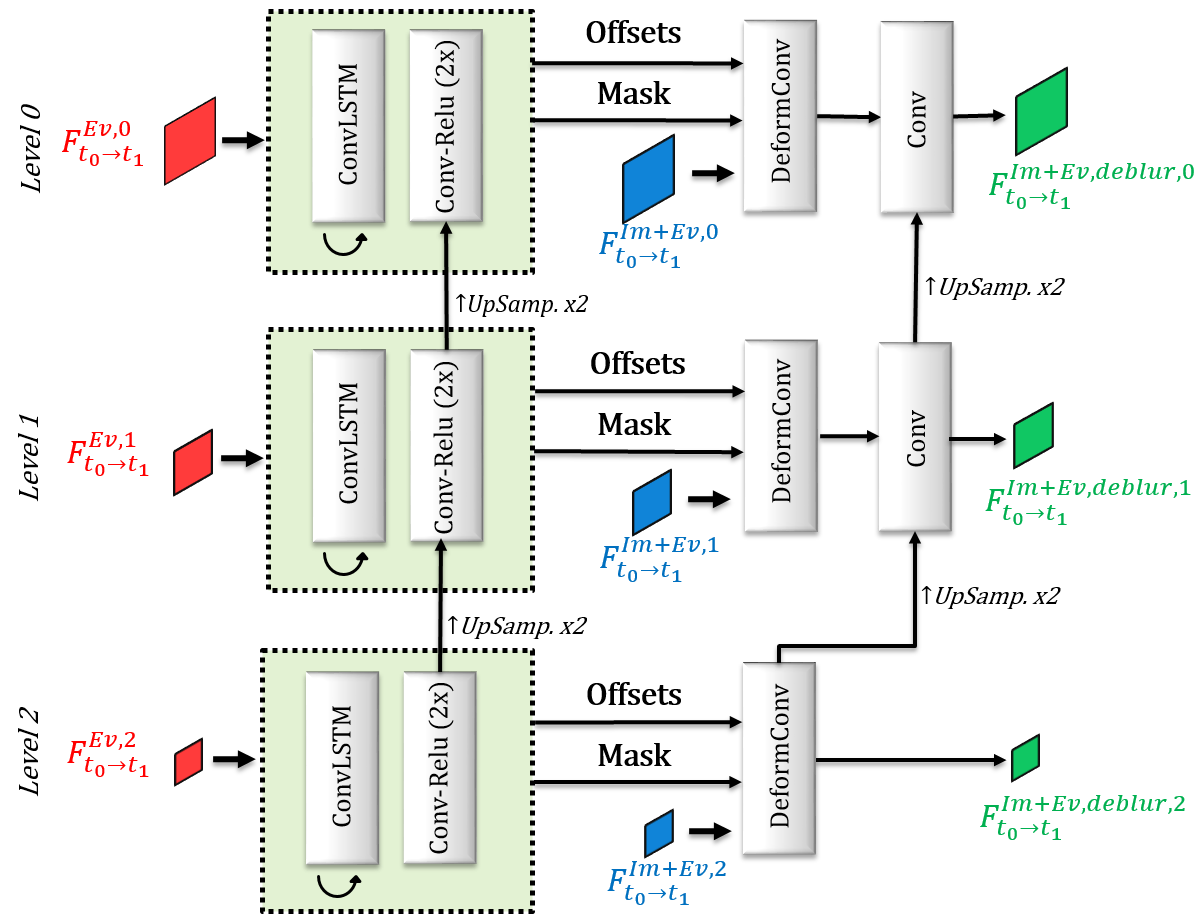}     & 
            \includegraphics[width=0.45\columnwidth]{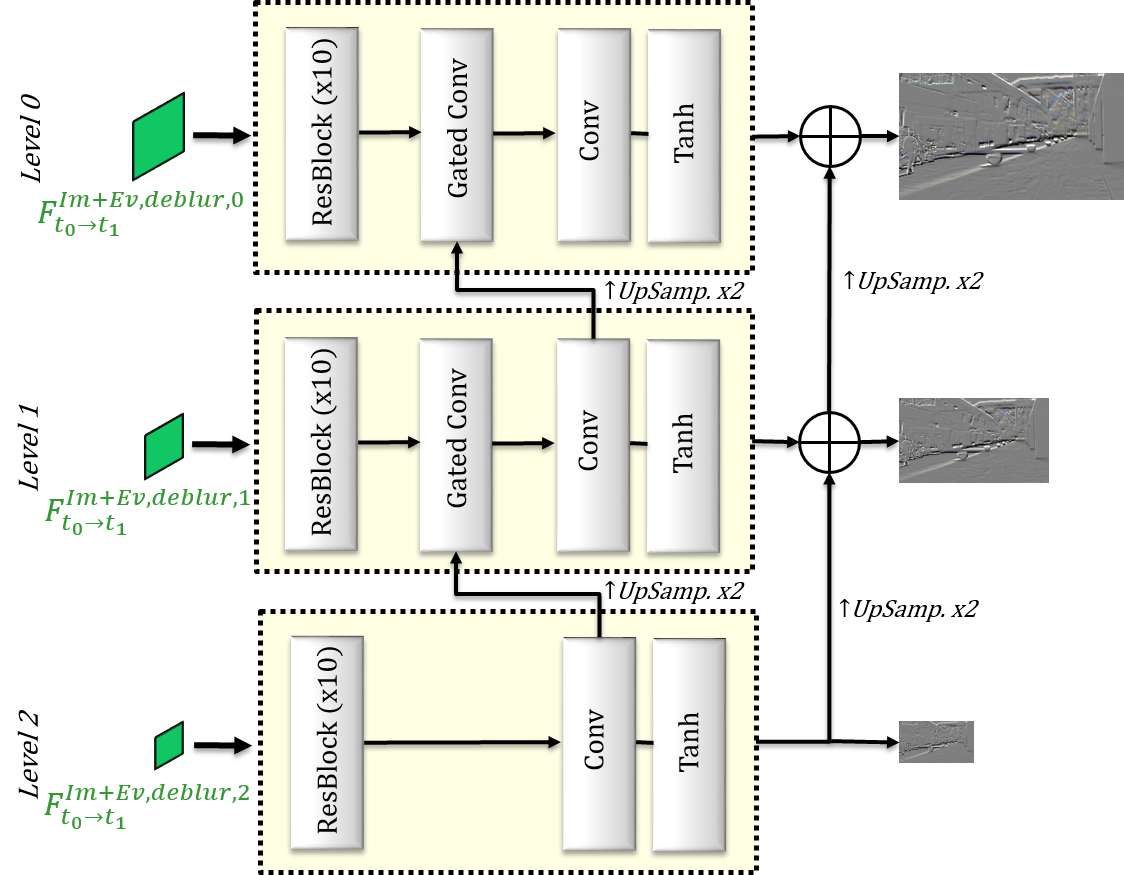}\\
   (b)  Deblur Module & (c) Coarse-to-fine Residual Decoder
\end{tabular}
    \caption{\textbf{Proposed Architecture.} (a)
    Two encoders (blue and red) are used to compute blurry image and events features $\{\textbf{F}^{Im+Ev}_{t_0 \rightarrow t_1}\}_{l=0}^L$ and events features $\{\textbf{F}^{Ev}_{t_0 \rightarrow t_1}\}_{l=0}^L$  that are used to compute the deblur operator and motion-related kernel offsets together with modulated masks, respectively.
    Then the (b) deblur module utilizes a set of modulated deformable convolutions to estimate the deblur $\{\textbf{F}^{Im+Ev,deblur}_{t_0 \rightarrow t_1}\}_{l=0}^L$ from the features $\{\textbf{F}^{Im+Ev}_{t_0 \rightarrow t_1}\}_{l=0}^L$ at different scales by using offsets and mask computed from $\{\textbf{F}^{Ev}_{t_0 \rightarrow t_1}\}_{l=0}^L$. The sharp image is decoded from the deblurred features by progressively reconstructing the residual image from coarser-to-finer scales.
    }
    \label{fig:abstract_overview}
\end{figure*}

To overcome the above-mentioned issues, in this work, we propose a divide-and-conquer deblurring approach that decouples the estimation of the per-pixel motion from the deblur operator. 
Due to their inherent temporal information, we use the input events to dynamically estimate per-pixel trajectories in terms of kernel offsets (and modulation masks) that correspond to the per-pixel motion.
On the other hand, the deblur operator is learned from a set of blurry images and corresponding event features and stays fixed after training.
This decoupling of tasks is achieved via the use of modulated deformable convolutions~\cite{zhu2019deformable}; see Fig.~\ref{fig:ilustrationDefConv} for an illustration.
Moreover, we frame our approach as a multi-scale, coarse-to-fine learning task to deal with the inherent sparsity of events.
In particular, coarser scales can take broader contextual information into account, while finer scales are encouraged to preserve the high-frequency details of the image, thereby alleviating the noisy and sparse nature of events. 
Notably, our algorithm works directly on high-resolution (HR) RGB images using only a monochrome event camera, thus exploiting the intra-correlation between channels and avoiding the intrinsic limitations of using color event cameras. 
An overview of our approach can be seen in Fig.~\ref{fig:abstract_overview}.

Due to the lack of real-world datasets containing HR RGB images with their corresponding events during the exposure time, and to demonstrate the ability of our method to generalize to real data, we collected a new dataset, called RGBlur+E, composed of real HR blurry images and events.

\textbf{Contributions} of this work are as follows
\begin{enumerate}[itemsep=0.25pt,topsep=0.25pt,parsep=0pt,partopsep=0pt]
    \item  We propose the first divide-and-conquer deblurring approach that decouples the estimation of per-pixel motion from the deblur operator. To explicitly exploit the dense motion information provided by the events, we employ modulated deformable convolutions. The latter allows us to estimate the integration locations (i.e. kernel offsets and modulation masks), which are solely related to motion, directly from events.
    
    \item To deal with the inherent sparsity of events as well as their lack in low contrast regions, we propose to employ a multi-scale, coarse-to-fine approach, that will exploit high frequency local details of events and broad motion information. 

    \item To evaluate and facilitate future research, we collect the first real-world dataset consisting of \emph{HR RGB Blur Images plus Events} (RGBlur+E). To allow both quantitative and qualitative comparisons, RGBlur+E consists of two subsets: the first contains sharp images, corresponding events, and synthetically generated blurry images to allow quantitative evaluations, and the second consists of real blurry images and corresponding events.  
    
    \end{enumerate}

\section{Related Work}

In this section, we discuss prior work on \textit{image deblurring}, i.e. given a blurry image as input (and optionally events during the exposure time), predict the sharp latent image at a certain timestamp of the exposure time. This is not to be mistaken with \textit{video deblurring}, i.e. from a blurry video (and optionally events) produce the sharp latent video, which is a different task (typically solved in conjunction with other tasks) not studied in this paper. 

\textbf{Image-based approaches.}
A generic image-based blur formation model can be defined as
\begin{equation}
      \textbf{B}(x,y)   
       =   \frac{1}{t_1-t_0}  \sum_{t=t_0}^{t_1} \textbf{I}_t(x,y) = \langle \mathbf{I}_{nn}(x,y), \mathbf{k}(x,y) \rangle  
\end{equation}
where $\textbf{B}\in \mathbb{R}^{H \times W \times C}$ is the blurry image over the exposure time $[t_0, t_1]$, $\textbf{I}_t\in \mathbb{R}^{H \times W \times C}$ is the sharp image at time $t\in[t_0, t_1]$, $\mathbf{I}_{nn}(x,y)$ is a window of size $K \times K$ around pixel $(x,y)$ and $\mathbf{k}$ is a per-pixel blur kernel.
Blind deblurring formulates the problem as one of blind deconvolution where the goal is to recover the sharp image $\textbf{I}_t$ without knowledge of the blur kernel $\mathbf{k}$~\cite{kundur1996blind}. 
Most traditional methods apply a two-step approach, namely blur kernel estimation followed by non-blind deconvolution~\cite{fergus2006removing,cho2009fast,pan2014deblurring,levin2009understanding}. 
Additionally, assumptions about the scene prior are made to solve the problem. 
Examples of such priors are the Total Variation (TV)~\cite{chan1998total}, sparse image priors~\cite{levin2009understanding,xu2013unnatural}, color channel statistics~\cite{pan2016blind,yan2017image}, patch recurrence~\cite{michaeli2014blind}, and outlier image signals~\cite{dong2017blind}. 
Nonetheless, most of those methods are limited to uniform blur. 
To handle spatially-varying blur, more flexible blur models have been proposed based on a projective motion path model~\cite{tai2010richardson,whyte2012non,zhang2013non}, or a segmentation of the image into areas with different types of blur~\cite{levin2006blind,dai2009removing,hyun2013dynamic,pan2016soft}.   
However, those methods are computationally heavy (e.g. computation time of 20 mins~\cite{sun2015learning} or an hour~\cite{hyun2013dynamic} per image) and successful results will depend on the accuracy of the blur models.

Image deblurring has benefited from the advances of deep convolutional neural networks (CNNs) by directly estimating the maping from blurry images to sharp. Several network designs have been proposed that increase the receptive field~\cite{zhang2018dynamic}, use multi-scale strategies~\cite{nah2017deep,noroozi2017motion,gao2019dynamic,tao2018scale,zhang2019deep,suin2020spatially,cho2021rethinking,chen2021hinet}, learn a latent presentation~\cite{tran2021explore,nimisha2017blur}, use deformable convolutions~\cite{yuan2020efficient,huo2021blind}, progressively refine the result~\cite{wieschollek2017learning,zamir2021multi,park2020multi}, or use perceptual metrics~\cite{kupyn2018deblurgan,kupyn2019deblurgan}. 
Also, some works combined model-based and learning-based strategies, by first estimating the motion kernel field using a CNN, and then applying a non-blind deconvolution~\cite{gong2017motion,sun2015learning,carbajal2021non}. 

However, despite advances in the field, image-based approaches struggle to reconstruct structure or perform accurate deblurring in the presence of large motions, also in real case scenarios and in the case of model-based approaches, they are bound by the employed model of the blur kernel.

\textbf{Event-based approaches.}
An event camera outputs a sequence of \emph{events}, denoted by $(x,y,t,\sigma)$, which record log intensity changes. $(x,y)$ are image coordinates, $t$ is the time the event occurred, and $\sigma=\pm 1$, called the polarity, denotes the sign of the intensity change. Due to their high temporal resolution, and hence reliably encoded motion information, event sensors have been used for image deblurring. 
Several event-based methods perform image deblurring on grayscale images, or alternatively on color images by processing color events channel-wise\footnote{Note that, the use of color events comes with its drawbacks. Small effective resolution and reduction of light per pixel, the need for demosaicking at the event level, and the lack of good commercial color event cameras, to mention a few.}. 
In~\cite{pan2019bringing}, the authors model the relationship between the blurry image, events, and latent frames through an Event-based Double Integral (EDI) model defined as,
\begin{equation}
    \textbf{B} =\frac{1}{t_1-t_0}  \sum_{t=t_0}^{t_1} \textbf{I}_t = \frac{\textbf{I}_{t_0}}{t_1-t_0} \odot  \sum_{t=t_0}^{t_1}  \exp{(c \textbf{E}_{t_0\rightarrow t})} 
\end{equation}
where $\textbf{E}_{t_0 \rightarrow t} \in \mathbb{R}^{H \times W}$ is the sum of events between $t_0$ and $t$, and $c$ is a constant. $\textbf{E}_{t_0 \rightarrow t}$ can be computed as the integral of events between $t_0$ and $t$.
However, the noisy and lossy nature of event data often results in strongly accumulated noise, and loss of details and contrast in the scene. 
In~\cite{pan2020high}, the authors extend the EDI model to the  multiple EDI (mEDI) model to get smoother results based on multiple images and their events.
\cite{jiang2020learning} combines the optimization approach of the EDI model with an end-to-end learning strategy. 
In particular, results obtained by~\cite{pan2019bringing} are refined by exploiting computed optical flow information and boundary guidance maps. 
Similarly,~\cite{wang2020event} proposes to incorporate an iterative optimization method into a deep neural network that performs denoising, deblurring, and super-resolution simultaneously.

Direct one-to-one mapping from a blurry image plus events has been performed by \cite{haoyu2020learning,zhang2020hybrid,xu2021motion,sun2021mefnet}.
\cite{haoyu2020learning} utilize two consecutive U-Nets to perform image deblurring followed by high frame rate video generation. 
In \cite{sun2021mefnet}, two encoders-based Unet are used where image and events features are combined using cross-modal attention and the skip connections across stages are guided by an events mask. Additionally, they propose an alternative representation to voxel grid that has into account the image deblurring formation process.
A drawback of \cite{haoyu2020learning,xu2021motion,sun2021mefnet} methods is that either the method just work on grayscale images or color events cameras are needed.
In~\cite{zhang2020hybrid}, the authors use the information of a monochrome event camera to compute an event representation through E2VID~\cite{rebecq2019events}. Then, the event representation is used as an additional input to an RGB image-based network~\cite{zhang2019deep}.
However, direct one-to-one mapping approaches often fail to generalize to real data. 
To overcome this issue, ~\cite{xu2021motion} use a self-supervised approach based on optical flow estimation where real-world events are exploited to alleviate the performance degradation caused by data inconsistency.

In this work, instead of learning all the deblurring process at once, we propose to decouple the learning of per-pixel motion from the deblur operators, and retain only the essential information for each sub-task. 
One of the  benefits of the proposed method is that while the implicit deblur operator is learnt and fixed, the per-pixel motion is dynamically learned for each single example adapting the sampling of the kernel in the deblur module to the motion of each specific blur image.
Moreover, to solve common artifacts of event-based methods, we use a multi-scale coarse-to-fine approach that handles the lack of events in low contrast regions and their noisy nature.  
Additionally, the proposed approach works directly with a combination of HR RGB images and monochrome event cameras and we propose the first dataset that combines HR RGB images with real events.

\begin{figure}[ht]
    \centering
    \begin{tabular}{cc}
        \includegraphics[width=0.2\columnwidth]{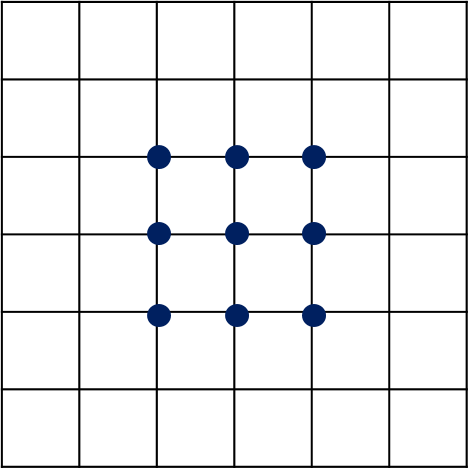}     &
                \includegraphics[width=0.2\columnwidth]{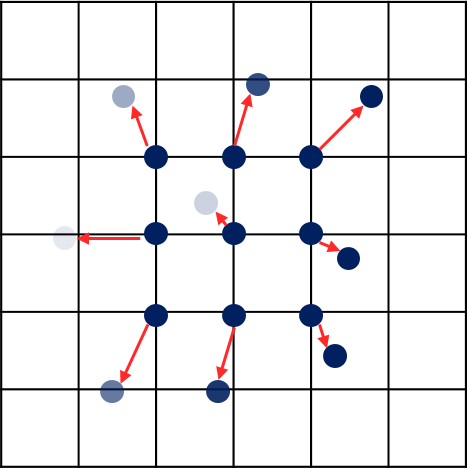}\\
      (a) Standard    & (b) Modulated Deformable   \\\
    \end{tabular}
    \caption{\textbf{Standard vs Deformable Convolution.} Illustration of kernel sampling locations in a $3\times3$ standard and modulated deformable convolution~\cite{zhu2019deformable}. (a) Regular sampling in a standard convolution. (b) In a modulated deformable convolution, the sampling deforms in the direction of the learnable offsets (red lines) and is weighted by the learnable modulation mask.}
    \label{fig:ilustrationDefConv}
\end{figure}

\textbf{Deformable convolutions.}
A major limitation of current (image- or event-based) deblurring methods is that by using standard convolutions, they end up applying the same deblur operators, learned during training, with a fixed receptive field across the entire image, regardless of the motion pattern that caused the blur. 
In contrast, deformable convolutions~\cite{dai2017deformable} learn kernel offsets per pixel, and thus can effectively change the sampling locations of the deblur operators.
This allows for more 'motion-aware' deblur operators with the ability to adapt their receptive field according to the motion pattern.
In~\cite{zhu2019deformable}, the authors propose to also learn a modulation mask in conjunction with kernel offsets, i.e. modulated deformable convolutions, to attenuate the contribution of each sampling location.
See Fig.~\ref{fig:ilustrationDefConv} for an illustration of the sampling process in a modulated deformable convolution.
In image deblurring, modulated deformable convolutions have been used to adapt the receptive field to the blur present in the scene~\cite{yuan2020efficient,huo2021blind}.
In this paper, we propose to use it differently. Since events contain fine-grained motion information per pixel, we can estimate 'motion-aware' deblur operators by conditioning the learning of kernel offsets and modulation masks solely to encoded event features. This allows the receptive field of deblur operators to dynamically adapt according to the underlying motion pattern.

\section{Method}

\textbf{Problem formulation}
Given a blurry RGB HR image $\mathbf{B}_{t_0 \rightarrow t_1}$ and the corresponding event stream $\textbf{E}_{t_0 \rightarrow t_1}$ containing all asynchronous events triggered during exposure time $[t_0, t_1]$, the proposed method reconstructs a sharp latent image $\mathbf{I}_{t}$ corresponding to the mid exposure timestamp $t=(t_1+t_0)/2$.

\subsection{System Overview}

Fig.~\ref{fig:abstract_overview} illustrates the proposed event-based image deblurring architecture, which consists of: an RGB blurry image and events encoder, a separate recurrent events encoder, a \emph{deblur} module, and a multi-scale residual decoder. 

The first encoder computes a multi-scale feature representation, $\{\textbf{F}^{Im+Ev}_{t_0 \rightarrow t_1}\}_{l=0}^L$, from the blurry image $\mathbf{B}_{t_0 \rightarrow t_1}$ and the stream of events $\textbf{E}_{t_0 \rightarrow t_1}$ during the exposure time represented as a \textit{voxel grid} with five equally-sized temporal bins. 

Similarly, the second encoder computes a multi-scale feature representation $\{\textbf{F}^{Ev}_{t_0 \rightarrow t_1}\}_{l=0}^L$, but solely from the events. Differently, the stream of events $\textbf{E}_{t_0 \rightarrow t_1}$ is split into fixed-duration chunks $\{\textbf{V}_{t_{(n-1)/N} \rightarrow t_{n/N}}\}_{n=1}^N$, with each chunk containing all events within a time window $1/N$ represented as a voxel grid with five equally-sized temporal bins. The event chunks are integrated over time using a ConvLSTM module.

\sloppy The deblur module exploits the features from the second encoder $\{\textbf{F}^{Ev}_{t_0 \rightarrow t_1}\}_{l=0}^L$ to dynamically estimate kernel offsets and modulation masks per pixel that adapt the sampling locations of the modulated deformable convolutions according to the per-pixel motion pattern of the input image. Sequentially, the modulated deformable convolutions are applied to the first encoder features $\{\textbf{F}^{Im+Ev}_{t_0 \rightarrow t_1}\}_{l=0}^L$, at each scale, essentially performing the deblur operators at each feature level.

Finally, the deblurred features $\{\textbf{F}^{Im+Ev,deblur}_{t_0 \rightarrow t_1}\}_{l=0}^L$ are passed through a multi-scale decoder that progressively reconstructs the residual image in a coarse-to-fine manner. 
By doing so, our method exploits both image- and event-based information.
Coarser scales will encode broad contextual information that helps deblurring if events are missing, while original resolution scales will estimate finer details by exploiting image and event structural information.

Sections~\ref{sec:DeblurMod} and~\ref{sec:Rec} detail the deblur module and coarse-to-fine reconstruction respectively.

\subsection{Deblur Module} \label{sec:DeblurMod}

A standard 2D convolution consists of two steps: 1) sampling on a regular grid $\mathcal{R}$ over the input feature map $\textbf{x}$; and 2) summation of sampled values weighted by $w$. 
The grid $\mathcal{R}$ defines the receptive field, and dilation.
As illustrated in Fig.~\ref{fig:ilustrationDefConv}(a), for deformable convolutions the regular grid $\mathcal{R}$ is augmented with offsets $\{\Delta x_k|n = 1, ..., N\}$, and in the case of modulated deformable convolutions additionally modulated scalars $\Delta m_k$ weight each position.

Blurry images by itself do not contain explicit temporal information or motion cues during the image acquisition, thus, being bad candidates as a clue to estimate kernel offsets and modulation masks. 
In contrast, due to their high temporal resolution, events contain all fine-grained motion cues during the exposure time that led to the blurry image, allowing for one-to-one mapping solutions.
Motivated by this observation, in this work, we exploit the inherent motion information in the event chunks to learn the offsets $\Delta x_k$ and modulated scalars $\Delta m_k$ that 1) adapt the sampling grid of the modulated deformable convolution to the per-pixel motion of the scene, and 2) weight every sampling position according to its actual contribution. 
The estimated offsets and masks are directly applied to $\textbf{F}^{Im+Ev}(x+x_k + \Delta x_k)$ to perform deblurring at the feature level. 
In particular, the modulated deformable convolutions can be expressed as: 
\begin{equation}
\textbf{F}^{Im+Ev,deblur}(x)= \sum_{k=1}^{K} w_k \cdot \textbf{F}^{Im+Ev}(x+x_k + \Delta x_k) \cdot \Delta m_k. 
\end{equation}
with $w_k$ being the kernel convolutional weights.
To address inherent limitations of event data, i.e. noisy events or lack of events in low-contrast regions, we perform the feature deblurring in a multi-scale fashion (see Section \ref{sec:Rec}. 
Specifically, as shown in Fig.~\ref{fig:abstract_overview}(b), to generate deblur features at the $l$-th scale we re-use the features computed at the $(l-1)$-th scale. 
Note that, in order to compute the offsets and mask we first integrate the chunk of events using a ConvLSTM module~\cite{xingjian2015convolutional} followed by two Conv+ReLU blocks that return the offsets and mask at each scale. 
The ConvLSTM layer will encode the overall temporal information and enable the network to learn the integration of events, as done analytically in the EDI formulation \cite{pan2019bringing} or recently is done similarly by the SCER preprocessing module in \cite{sun2021mefnet}.

\subsection{Multi-scale Coarse-to-Fine Approach} \label{sec:Rec}

By nature, event information is inherently sparse, noisy and they lack in low contrast regions.  
To address this problem, we propose to, first, perform feature deblurring in a multi-scale fashion and, second, progressively estimate the sharp image by computing residuals at different scales from multi-scale deblur features $\{\textbf{F}^{Im+Ev,deblur}_{t_0 \rightarrow t_1}\}_{l=0}^L$.

Coarse-to-fine reconstruction has been widely used in image-based restoration methods \cite{wieschollek2017learning,zamir2021multi,park2020multi}. While the motivation of those methods is to recover the clear image progressively by diving into the problem in small sub-tasks, in this work, the goal of this approach is to deal with the sparsity and lack of events in some regions. 
Coarser scales will learn broader contextual information where motion can be estimated even in locations where there are no events and being then more robust to noisy or missing events in low-contrast areas.  On the other hand, scales closer to the original resolution will exploit the fine details contained in the event data translating to details preservation in the final output image.

Specifically, as shown in Fig.~\ref{fig:abstract_overview}(b-c),  deblurred features are estimated at different scales $\{\textbf{F}^{Im+Ev, deblur}_{t_0 \rightarrow t_1}\}_{l=0}^L$ as explained in Section~\ref{sec:DeblurMod} and used to perform a coarse-to-fine residual reconstruction using an individual decoder block at each scale.
In particular, we employ a decoder at each scale composed of ten residual blocks, each block consisting of Conv+ReLU+Conv, an \emph{attention mechanism} (except in the last scale) that combines features from the current scale $l$ and the previous scale $l-1$, and a convolutional layer. The attention mechanism uses gating to select the most informative features from each scale.
Note that, the residual estimation at scale $l$ will be added to the estimation at the previous scale $l-1$.

\subsection{Loss Function}
To train the network, a sum over all the scales $l$ of a combination of an $L1$ image reconstruction loss  $L_{rec}^l$ , and a perceptual loss~\cite{zhang2018perceptual} $L_{percep}^l$ is used.

\noindent\textbf{Reconstruction loss.} The $L1$ reconstruction loss at a level $l$ is defined as
\begin{equation}
        L_{rec}^l =  || \textbf{I}\downarrow_{2l} - \tilde{\textbf{I}}_l ||_{1} 
\end{equation}
where $\textbf{I}\downarrow_{2l}$ denotes the sharp image downsampled by a factor of $N$, and $\tilde{\textbf{I}}_l$ the estimated sharp image equals to the sum of the blur image $\mathbf{B}_{t_0 \rightarrow t_1}\downarrow_{2l}$ and the learned residual $\mathbf{R}_{l}$.

\noindent\textbf{Perceptual loss.} The perceptual loss computes a feature representation of the reconstructed image and the target image by using a VGG network \cite{simonyan2014very} that was trained on ImageNet \cite{russakovsky2015imagenet}, and averages the distances between VGG features across multiple layers as follows
\begin{equation}
    L_{percep}^l =  \sum_n\frac{1}{H_nW_n}\sum_{h,w}\|\omega_n\odot (\Phi_n(\textbf{I}\downarrow_{2l})_{h,w} 
     - \Phi_n(\tilde{\textbf{I}}_l)_{h,w} )\|_2^2 ,
\label{eq:LPIPS}
\end{equation}
where $H_l$ and $W_l$ are the height and width of feature map  $\Phi_n$ at layer $n$ and $\omega_n$ are weights for each features. Note that, features are unit-normalized in the channel dimension.  
By using the perceptual loss (see Eq.~\ref{eq:LPIPS}), our network effectively learns to endow the reconstructed images with natural statistics (i.e. with features close to those of natural images).

\section{RGBlur+E Dataset}

Due to a lack of datasets for evaluating event-based deblurring algorithms on RGB HR images, we capture the first dataset containing real RGB blur images together with the corresponding events during exposure time, called \emph{RGBlur+E}.

\begin{figure}
\begin{tabular}{cc}
\includegraphics[height= 0.17\textheight]{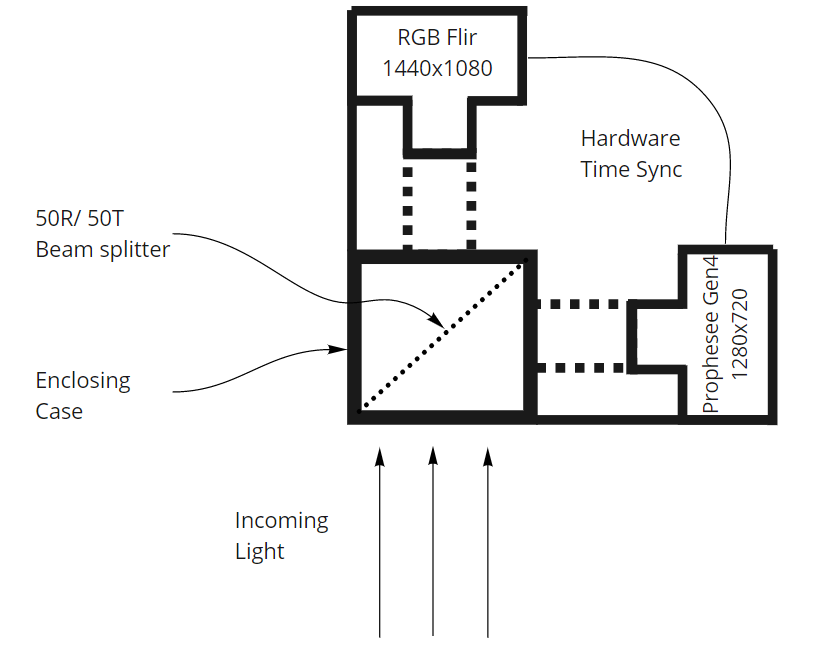}     &  
\includegraphics[height= 0.17\textheight]{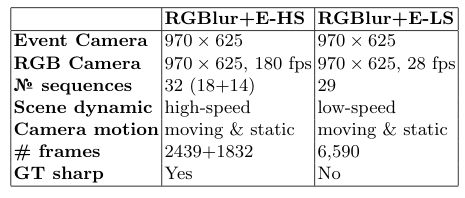} \\
(a) Beamsplitter setup & (b) Dataset Specifications

\end{tabular}

    \label{tab:datasetSpec}
    \caption{\textbf{RGBlur+E setup and specifications.} Image acquisition diagram and specifications summary of the proposed RGBlur+E dataset.}
\end{figure}

As shown in Tab.~\ref{tab:datasetSpec}a, we built a hybrid setup that uses a high-speed FLIR BlackFly S RGB GS camera 
and a Prophesee Gen4 monochrome event camera mounted on a rigid case with a 50/50 one-way mirror to share incoming light. The final resolution of both image and event data is $970 \times 625$. Both have been hardware synchronized and built into a beamsplitter setup. We have recorded a wide range of scenarios (cars, text, crossroads, people, close objects, buildings) and covered different motion patterns (static, pan, rotation, walk, hand-held camera, 3D motion). More details are provided in Tab.~\ref{tab:datasetSpec}b and the Supp. Mat. 

The dataset is composed by two subsets: \emph{RGBlur+E-HS} and \emph{RGBlur+E-LS}.
In RGBlur+E-HS, we use high-speed video sequences at $180$ fps to generate synthetic blurry images by averaging a fixed number of frames with ground truth sharp images and events during the exposure time.  
We collect a total of $32$ video sequences. We split the dataset into $18$ sequences for training and $14$ for testing. After synthesizing the blur images, the dataset contains 7600 pairs or frames for training and 3586 for testing.
In RGBlur+E-LS, we recorded $29$ sequences at $28$ fps with a total of 6590 frames with a longer exposure time containing a combination of blurry images and events during the exposure time. However, there is no ground truth data available.

\section{Experiments}

\subsection{Experimental Settings}

\textbf{Data preparation.}

\textbf{\emph{GoPro.}}
The GoPro dataset is widely adopted for image- and event-based deblurring.
To synthesize blurry images we average between 7 and 13 frames for training. For testing, we fix the number of frames to 11. Events are synthesized using the ESIM simulator~\cite{rebecq2018esim}.

\textbf{\emph{RGBlur+E dataset.}} To close the gap between simulated and real events, we finetune the pretrained model on the GoPro on the training set of the training subset of the proposed RGBlur+E-HS.

\textbf{Training setting.}
All experiments in this work are done using the PyTorch framework. For training, we use the Adam optimizer\cite{kingma2014adam} with standard settings, batches of size $6$ and learning rate $10^{-4}$ decreased by a factor $2$ every $10$ epoch. 
We train the full pipeline for $50$ epochs. The total number of parameters to optimize is $5.7$M with an inference time of $1.8$ sec. Notice the reduced number of parameters compared to other state of the art methods.

\subsection{Ablation of the effectiveness of the components}

Table~\ref{tab:ablationStudy} show an ablation study that analyzes the contribution of each component on the final performance of the algorithm. 
We see, (1) that event data eases the problem of image deblurring by boosting the performance by +3.18dB,
(2) that adapting the receptive field of the convolutions by using a second encoder to compute offsets and masks in the deblur module helps in the performance of the algorithm (+1.18 dB)
(3) that pre-processing the event by using a ConvLSTM is more informative for image deblurring than just using the voxel grid representation  (+1.69dB) 
(4) the benefits coarse-to-fine approach rather than computing the residual image directly at the finer scale (+0.29dB).

\begin{table}[h!] 
 \setlength{\tabcolsep}{1.1pt}
  \centering
  \begin{tabular}{@{}ccccc|ccc@{}}

    Im. &  C2F &  Events & DM & LSTM & PSNR $\uparrow$ & SSIM $\uparrow$  & LPIPS $\downarrow$  \\
    \hline
    \hline
        \checkmark & &  & &  &  28.28& 0.86 & 0.19 \\ 
    \checkmark &\checkmark   & & &  & 28.57  & 0.87 & 0.18\\ 
\checkmark & \checkmark & \checkmark  & &   & 31.46 & 0.92 & 0.15 \\
        \checkmark & \checkmark & \checkmark  &  \checkmark  &   & 32.64 & 0.93  & 0.13 \\
    \hline
\checkmark& \checkmark &  \checkmark &\checkmark &  \checkmark & \textbf{34.33} &\textbf{0.94}  & \textbf{0.11}  \\
  \end{tabular}
  \caption{\textbf{Components analysis on the GoPro dataset  \cite{nah2017deep}.}  Im. stands for image input, Events for their use as an additional input, DM for Deblur Module computed from events and C2F for coarse-to-fine reconstruction. }
  \label{tab:ablationStudy}
\end{table}

\begin{table}[h!bt]
    \centering
    \scriptsize
    \begin{tabular}{lccc}
\textbf{Method} & \textbf{Events} & \textbf{PSNR}  $\uparrow$  & \textbf{SSIM}  $\uparrow$ \\
\hline
\textbf{Model-based} \\
\hline
Xu et al. \cite{xu2013unnatural}  & \xmark  & \textcolor{gray}{21.00}& \textcolor{gray}{0.741} \\
 Hyun et al. \cite{hyun2013dynamic}  & \xmark & \textcolor{gray}{23.64} & \textcolor{gray}{0.824} \\
 Whyte et al. \cite{whyte2012non}   & \xmark & \textcolor{gray}{24.60} & \textcolor{gray}{0.846} \\
 Gong et al. \cite{gong2017motion}   & \xmark & \textcolor{gray}{26.40} & \textcolor{gray}{0.863}  \\
Carbajal et al. \cite{carbajal2021non}   & \xmark &  \textcolor{gray}{28.39} & \textcolor{gray}{0.82} \\
 \hline 
 \textbf{Learning-based} \\
 \hline
  Jin et al. \cite{jin2018learning}  & \xmark &  \textcolor{gray}{26.98}  & \textcolor{gray}{0.892}   \\
   Kupyn et al. \cite{kupyn2018deblurgan}     & \xmark & \textcolor{gray}{28.70} & \textcolor{gray}{0.858}  \\
 Nah et al. \cite{nah2017deep}    & \xmark &  \textcolor{gray}{29.08}  & \textcolor{gray}{0.913}  \\
 Zhang et al. \cite{zhang2018dynamic}    & \xmark &  \textcolor{gray}{29.18}    & \textcolor{gray}{0.931}  \\
 Kupyn et al. v2 \cite{kupyn2019deblurgan}    & \xmark & \textcolor{gray}{29.55}  &  \textcolor{gray}{0.934} \\
 Yuan et al.  \cite{yuan2020efficient}  & \xmark  &  \textcolor{gray}{29.81} & \textcolor{gray}{0.937}   \\
 Zhang et al. \cite{zhang2019deep}  &  \xmark & 30.23   &  0.900 \\
 Tao et al. \cite{tao2018scale}     & \xmark & 30.27 & 0.902   \\
Gao et al. \cite{gao2019dynamic} & \xmark & 31.17  & 0.916 \\
Park et al. \cite{park2020multi}    & \xmark  &  31.15  & 0.916 \\
Suin et al. \cite{suin2020spatially} &  \xmark  &  \textcolor{gray}{31.85} & \textcolor{gray}{0.948} \\
Huo et al. \cite{huo2021blind}  &  \xmark &  32.09  & 0.931 \\
Cho et al. \cite{cho2021rethinking} & \xmark & 32.45 & 0.934 \\
Zamir et al. \cite{zamir2021multi}  &  \xmark &  32.65  & 0.937\\ 
Chen et al. \cite{chen2021hinet} & \xmark & 32.76 & 0.937  \\
\hline 
 \textbf{Event-based} \\
 \hline
Pan et al. \cite{pan2019bringing}  &   Color &  \textcolor{gray}{29.06} & \textcolor{gray}{0.943} \\
Pan et al. \cite{pan2020high} &  Color & \textcolor{gray}{30.29} & \textcolor{gray}{0.919} \\ 
Jiang et al. \cite{jiang2020learning}   & Color  & \textcolor{gray}{31.79} & \textcolor{gray}{0.949} \\
Chen et al. \cite{haoyu2020learning}$^1$ &  Color & \textcolor{gray}{32.99}$^1$  & \textcolor{gray}{30.935}$^1$ \\
Zhang et al. \cite{zhang2020hybrid} &  Gray &  \textcolor{gray}{32.25} & \textcolor{gray}{0.929} \\
Ours & Gray   & \textbf{34.33} & \textbf{0.944} \\ 
\hline
    \end{tabular}
    \caption{\textbf{Quantitative comparisons on the synthetic GoPro dataset \cite{nah2017deep}.} Chen et al. ($^1$) computes the values by averaging 7 frames instead of 11. Gray values are extracted directly from the paper. 
    }
    \label{tab:resultsGoPro}
\end{table}

\begin{figure*}[!ht]
\centering
\setlength\tabcolsep{1.5pt}

\begin{tabular}{cccc}

\includegraphics[width=0.18\textwidth]{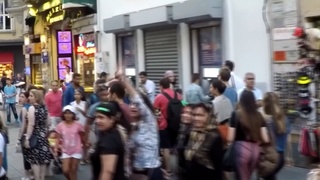} &
\includegraphics[width=0.18\textwidth]{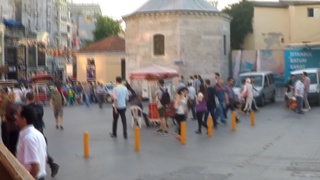} & 
\includegraphics[width=0.18\textwidth]{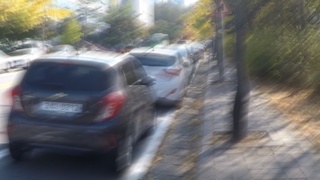}
 & 
\includegraphics[width=0.18\textwidth]{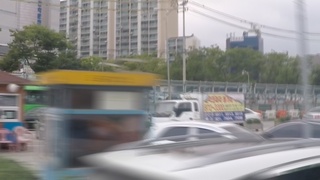}
\\
\end{tabular}
\begin{tabular}{cccccc}
 Huo et al. \cite{huo2021blind} &  Cho et al. \cite{cho2021rethinking} &  Chen et al. \cite{chen2021hinet}  & Zamir et al. \cite{zamir2021multi} & Pan et al.   \cite{pan2019bringing} &  Ours \\
 
\includegraphics[width=0.15\textwidth, trim=200 150 350 100,clip]{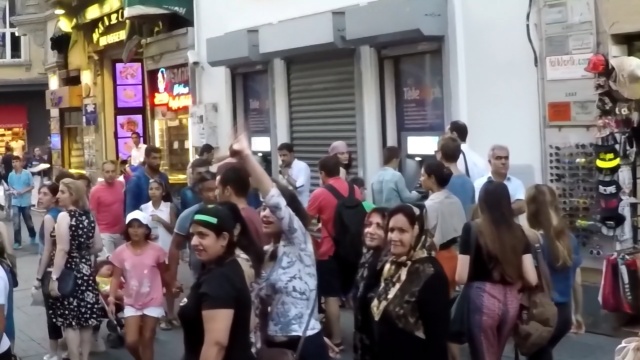}&
\includegraphics[width=0.15\textwidth, trim=200 150 350 100,clip]{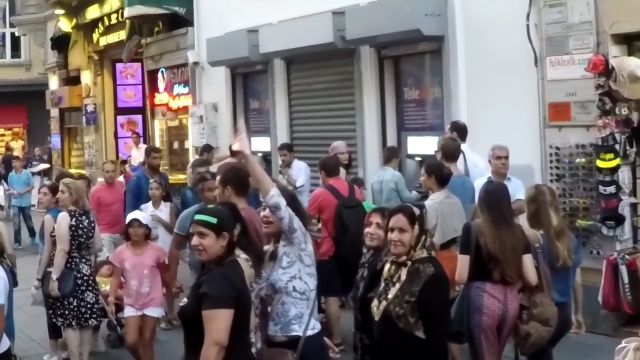} & 
\includegraphics[width=0.15\textwidth, trim=200 150 350 100,clip]{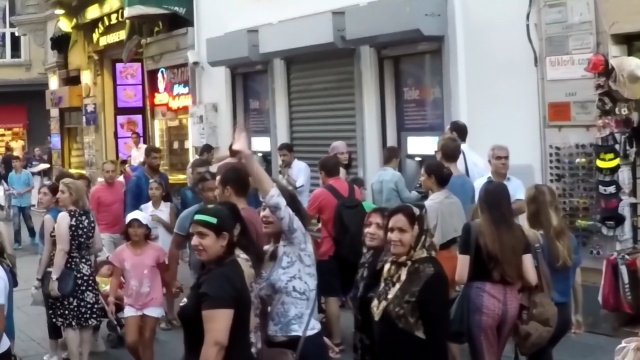}&
\includegraphics[width=0.15\textwidth, trim=200 150 350 100,clip]{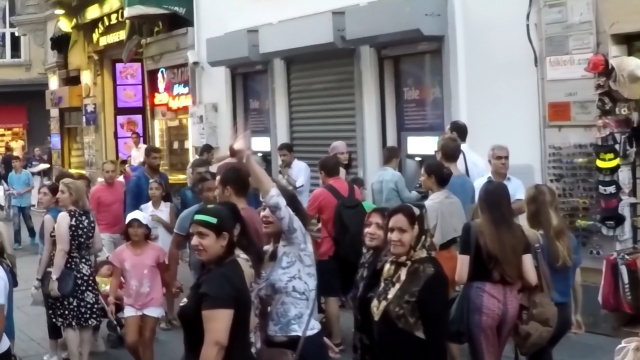} & 
\includegraphics[width=0.15\textwidth, trim=200 150 350 100,clip]{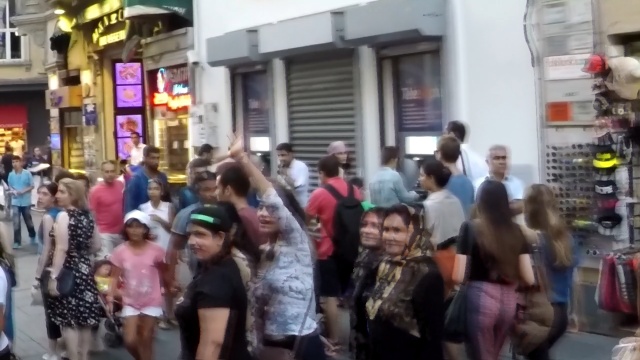} & 
\includegraphics[width=0.15\textwidth, trim=200 150 350 100,clip]{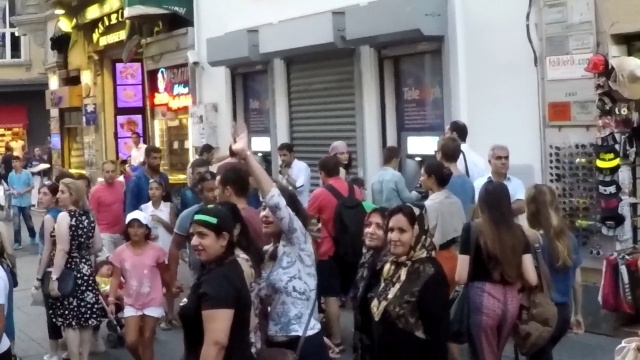} \\

\includegraphics[width=0.15\textwidth, trim=480 120 400 360,clip]{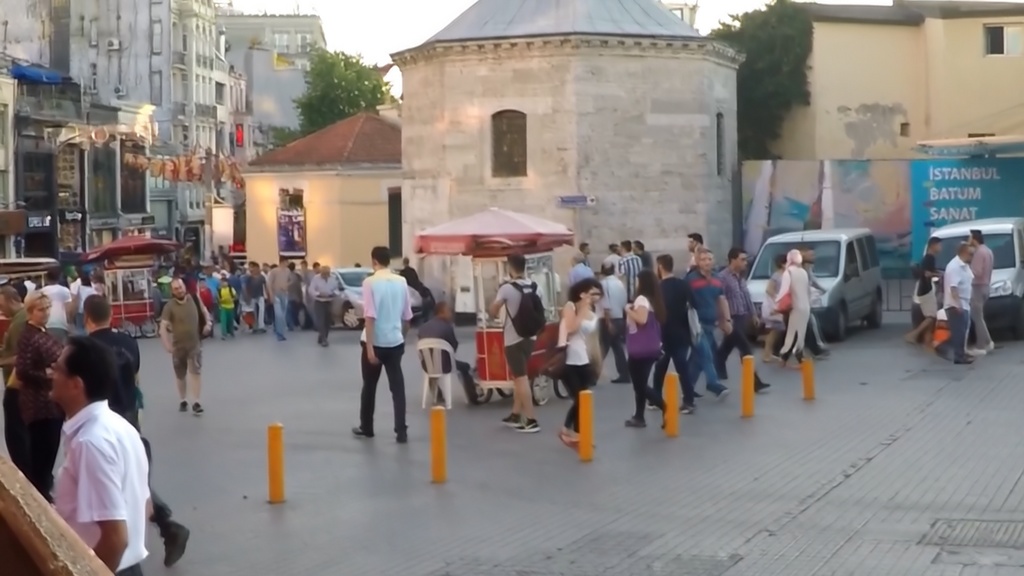} & 
\includegraphics[width=0.15\textwidth, trim=480 120 400 360,clip]{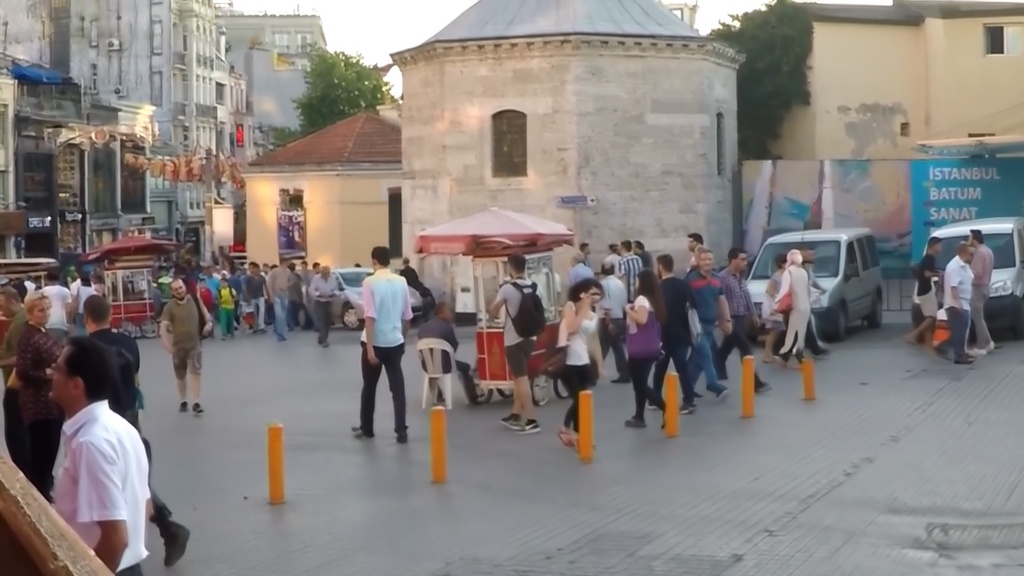} &
\includegraphics[width=0.15\textwidth, trim=480 120 400 360,clip]{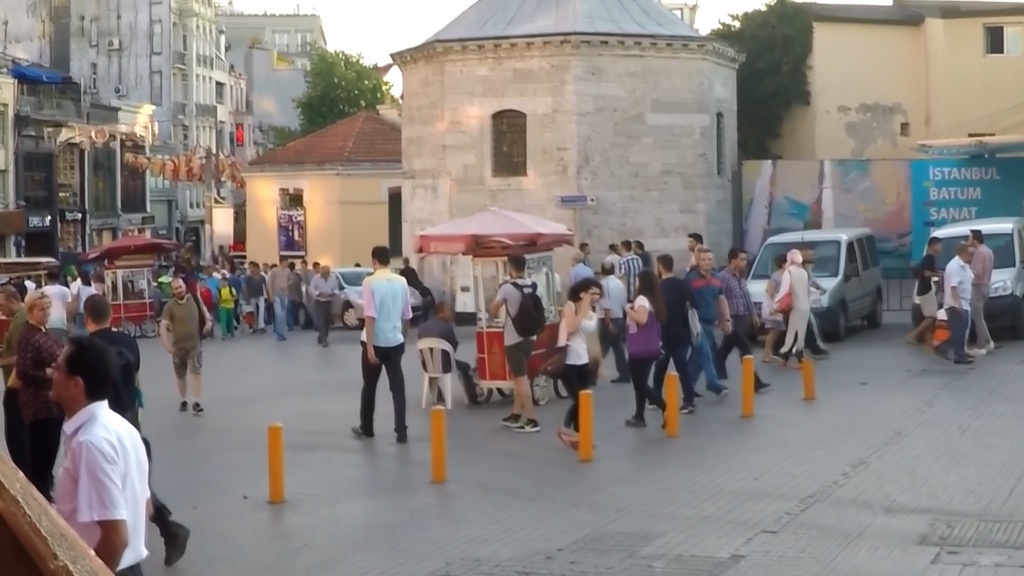} & 
\includegraphics[width=0.15\textwidth, trim=480 120 400 360,clip]{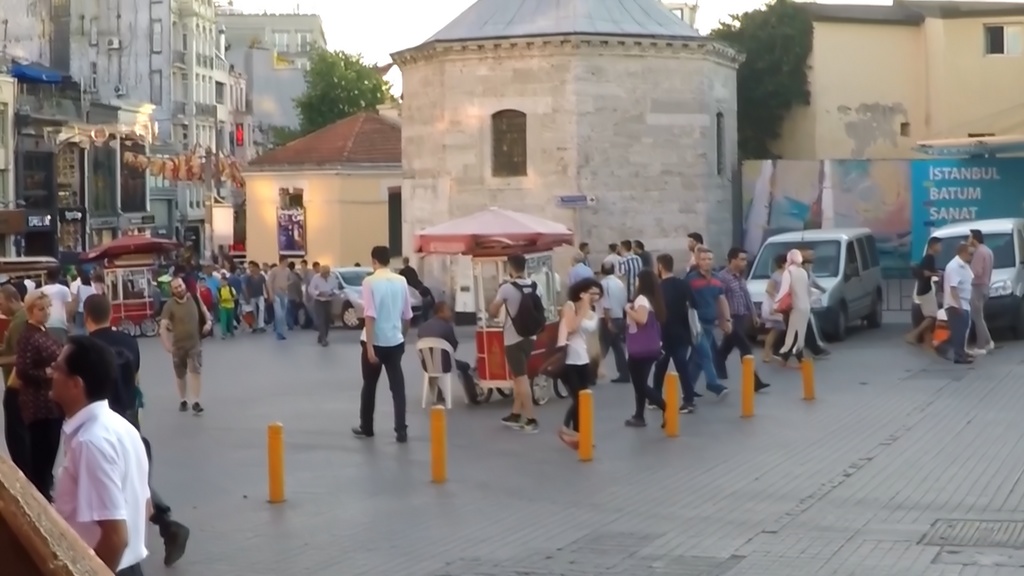} &
\includegraphics[width=0.15\textwidth, trim=480 120 400 360,clip]{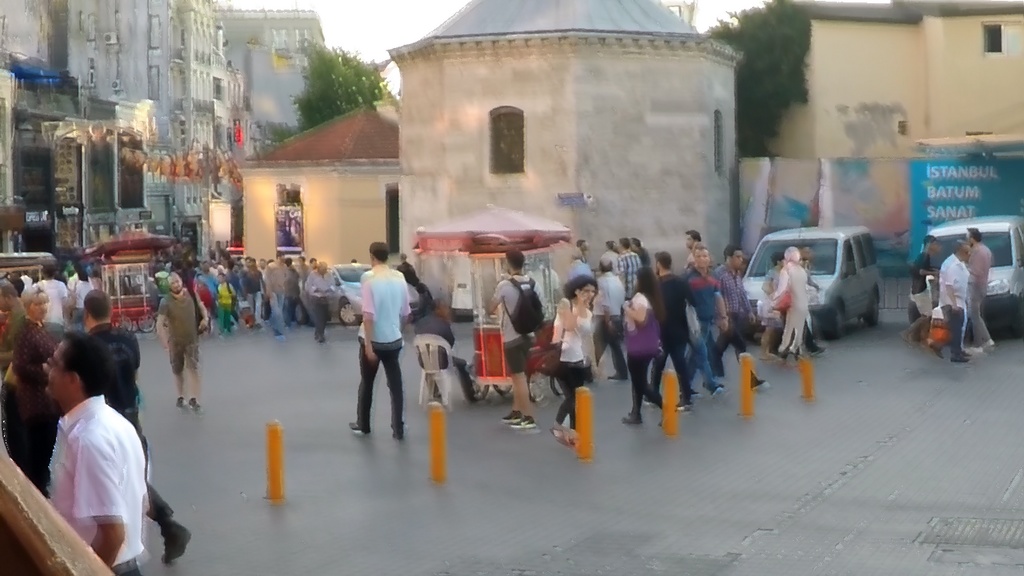}& 
\includegraphics[width=0.15\textwidth,  trim=480 120 400 360,clip]{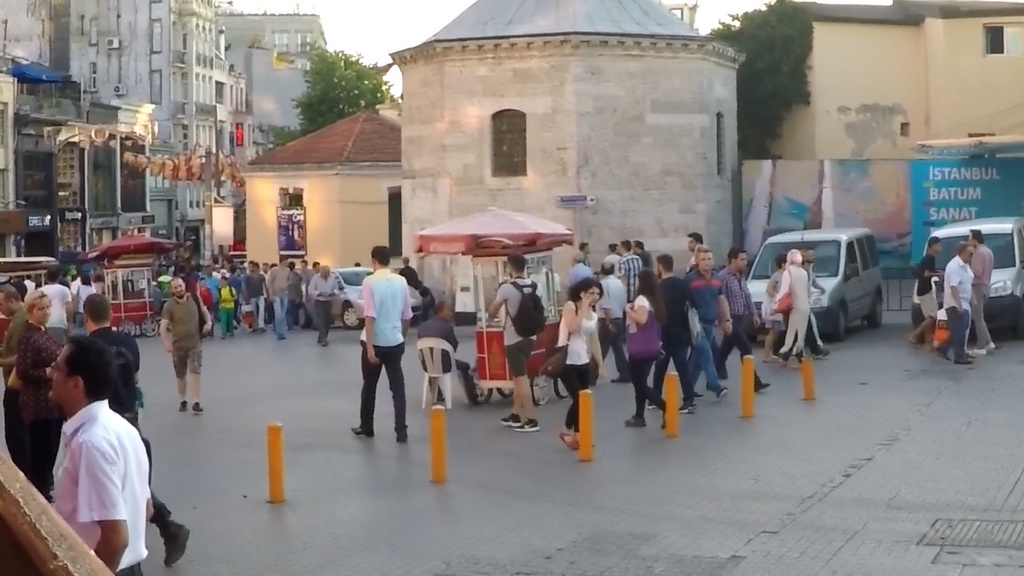}\\ 

\includegraphics[width=0.15\textwidth, trim=520 200 400 280,clip]{images/Results/GoPro/ASPDC/34_530_29.008771896362305_compressed.jpeg}& 
\includegraphics[width=0.15\textwidth, trim=520 200 400 280,clip]{images/Results/GoPro/MIMO/34_530_compressed.jpeg} &
\includegraphics[width=0.15\textwidth, trim=520 200 400 280,clip]{images/Results/GoPro/hinet/34_530_compressed.jpeg}& 
\includegraphics[width=0.15\textwidth, trim=520 200 400 280,clip]{images/Results/GoPro/MPRNet/34_530_32.317691802978516_compressed.jpeg} &
\includegraphics[width=0.15\textwidth, trim=520 200 400 280,clip]{images/Results/GoPro/pan_rgb/004_compresed.jpeg} & 
\includegraphics[width=0.15\textwidth,  trim=520 200 400 280,clip]{images/Results/GoPro/ours_ECCV/img_00000003_530_psnr_+34.673866271972656_pred_compressed.jpeg} \\

\includegraphics[width=0.15\textwidth, trim=150 200 900 400,clip]{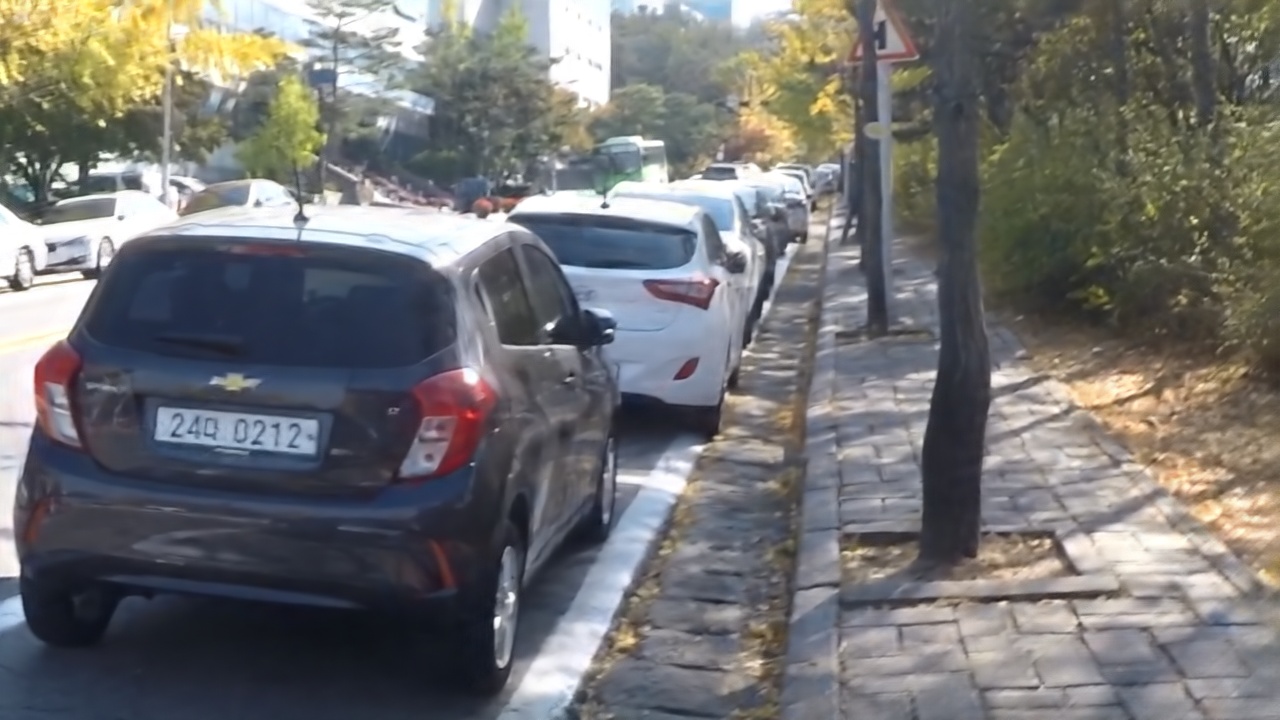} &
\includegraphics[width=0.15\textwidth, trim=150 200 900 400,clip]{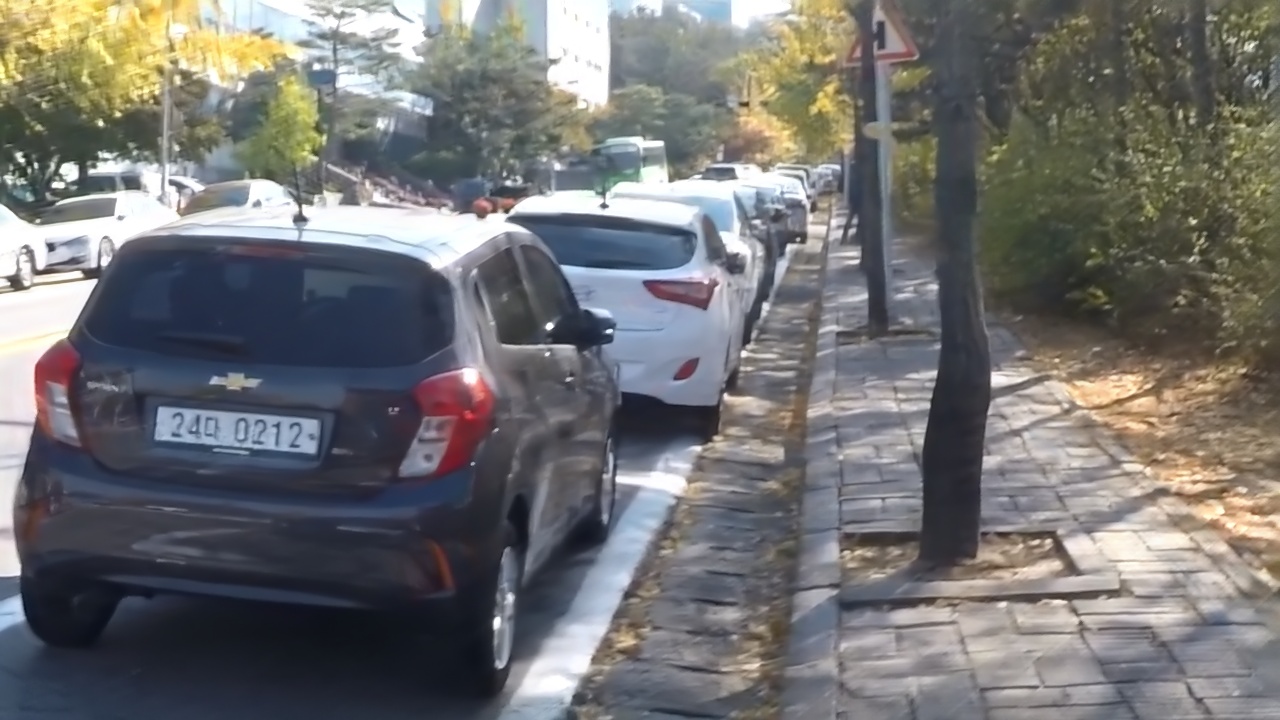}&
\includegraphics[width=0.15\textwidth,  trim=150 200 900 400,clip]{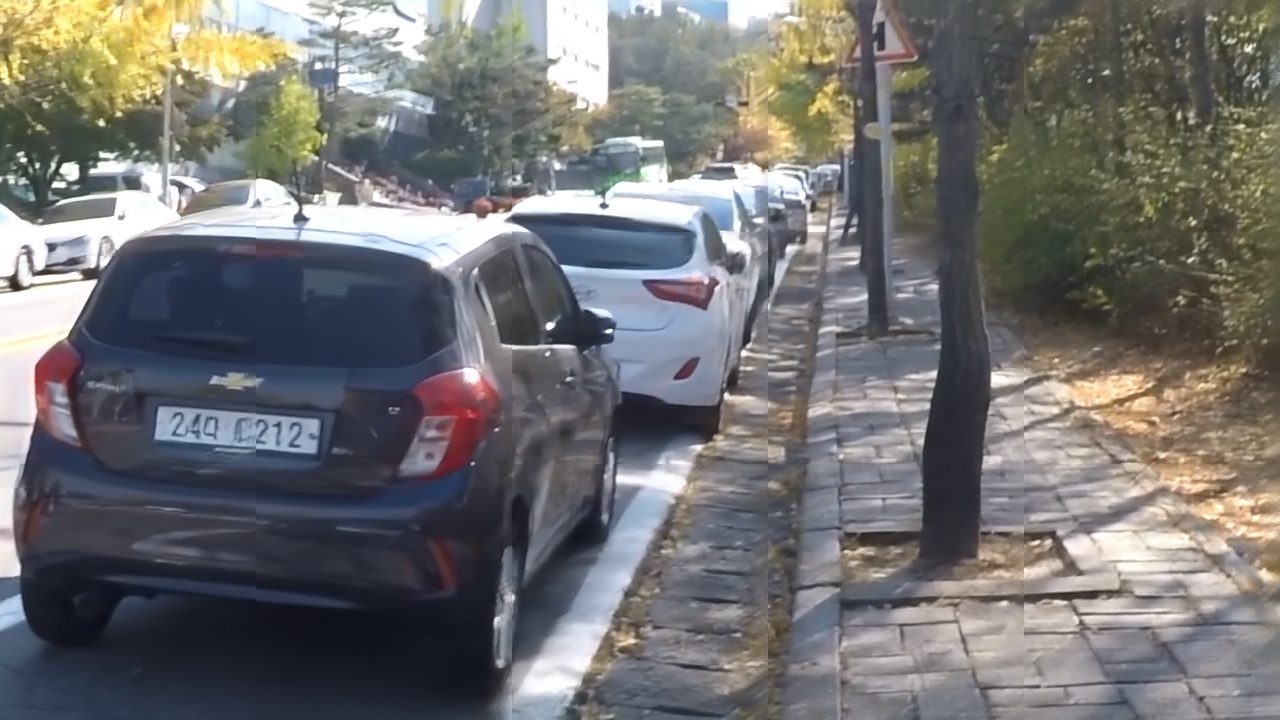}& 
\includegraphics[width=0.15\textwidth,  trim=150 200 900 400,clip]{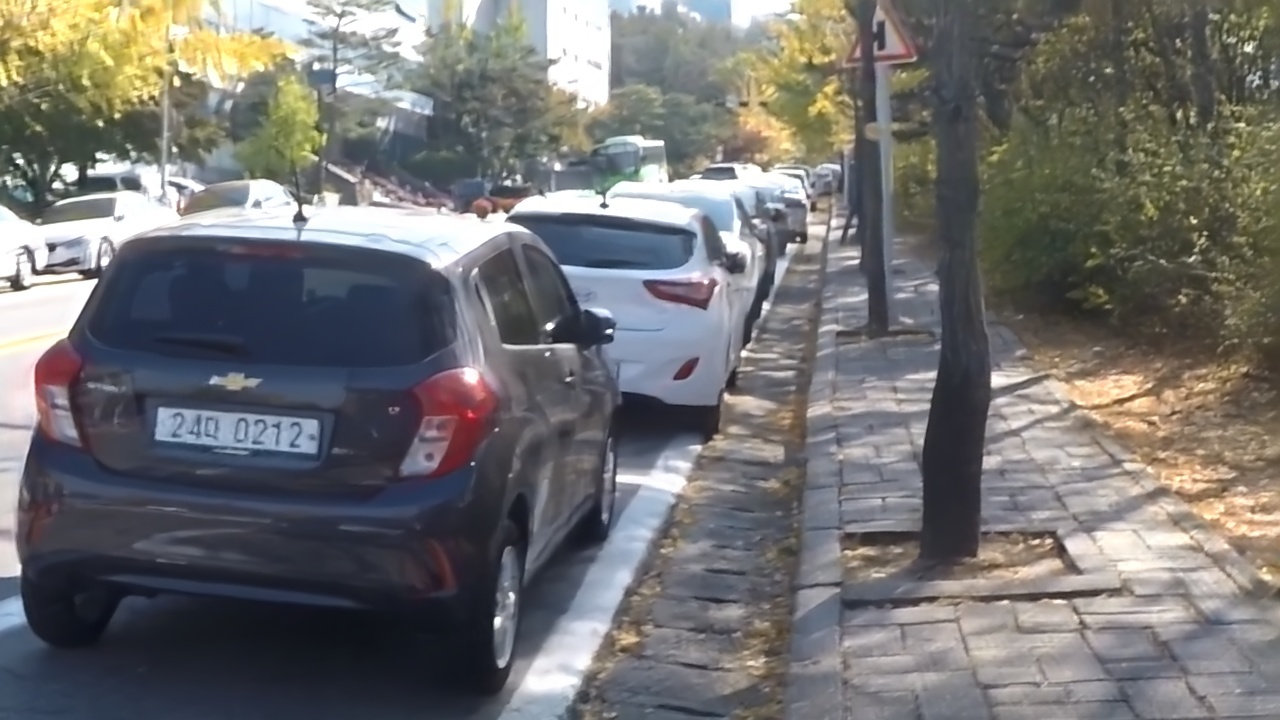} & 
\includegraphics[width=0.15\textwidth, trim=150 200 900 400,clip]{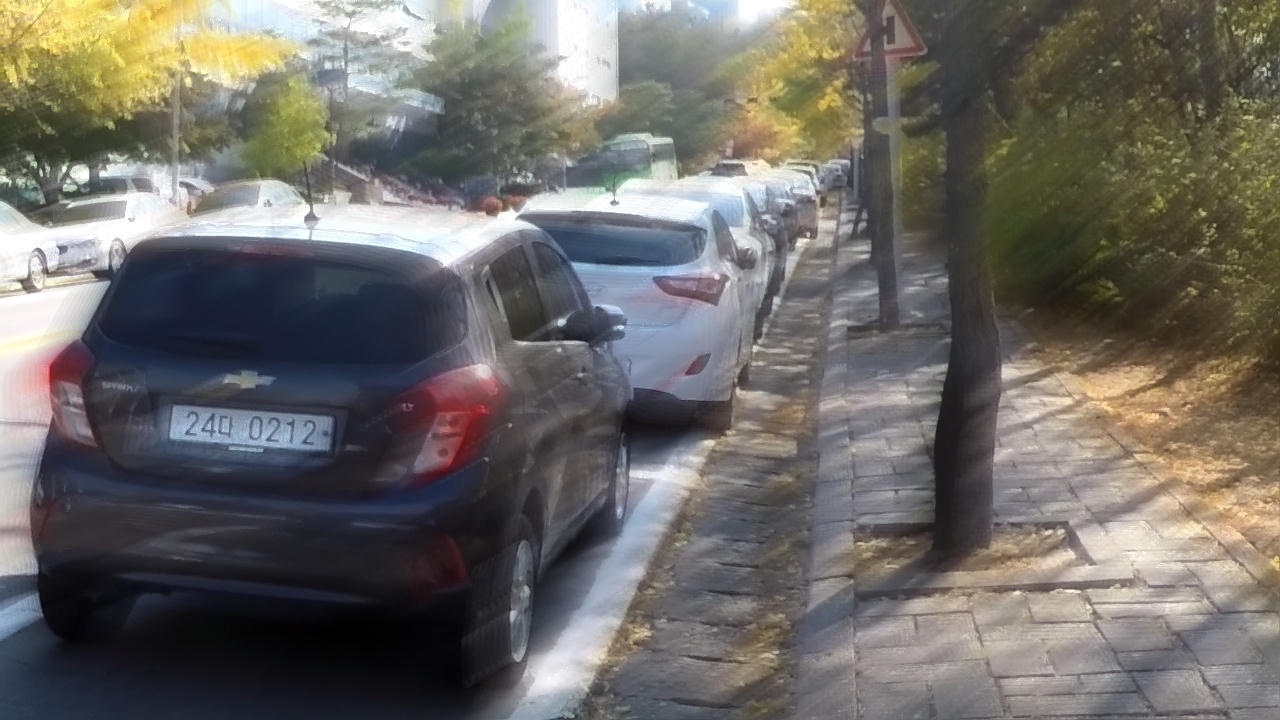} & 
\includegraphics[width=0.15\textwidth, trim=150 200 900 400,clip]{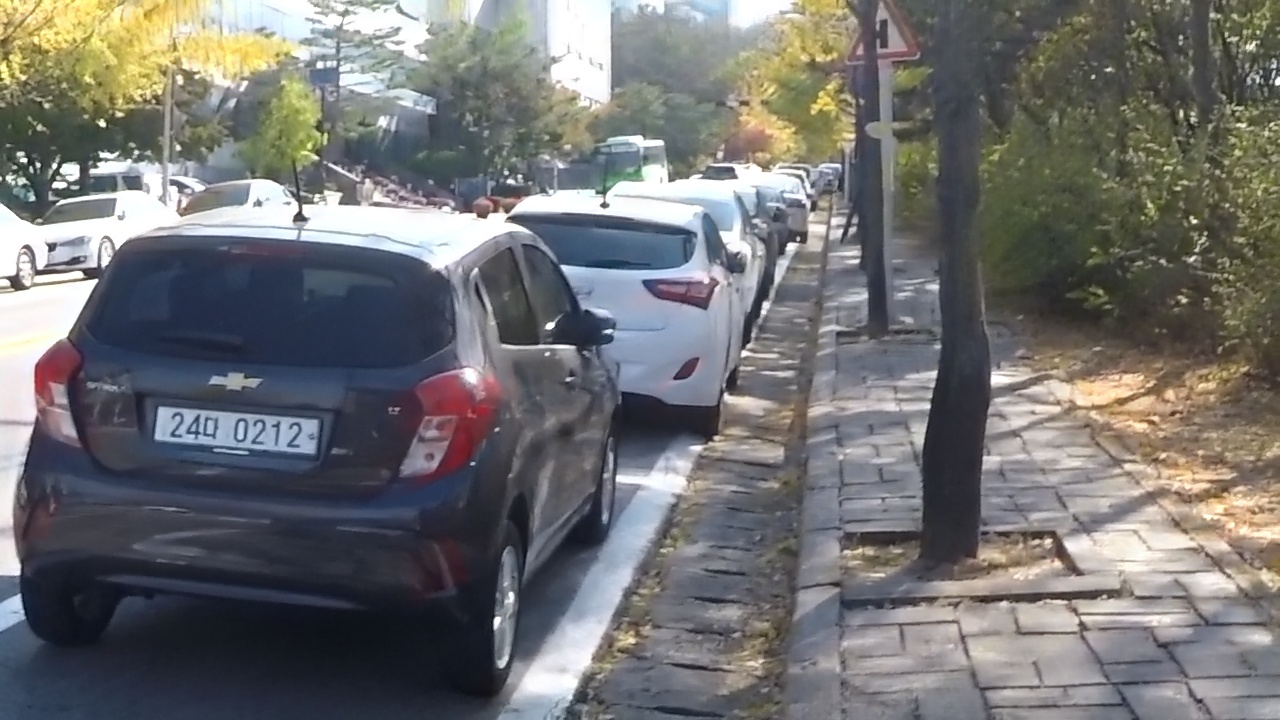} \\

\includegraphics[width=0.15\textwidth, trim=200 300 700 200,clip]{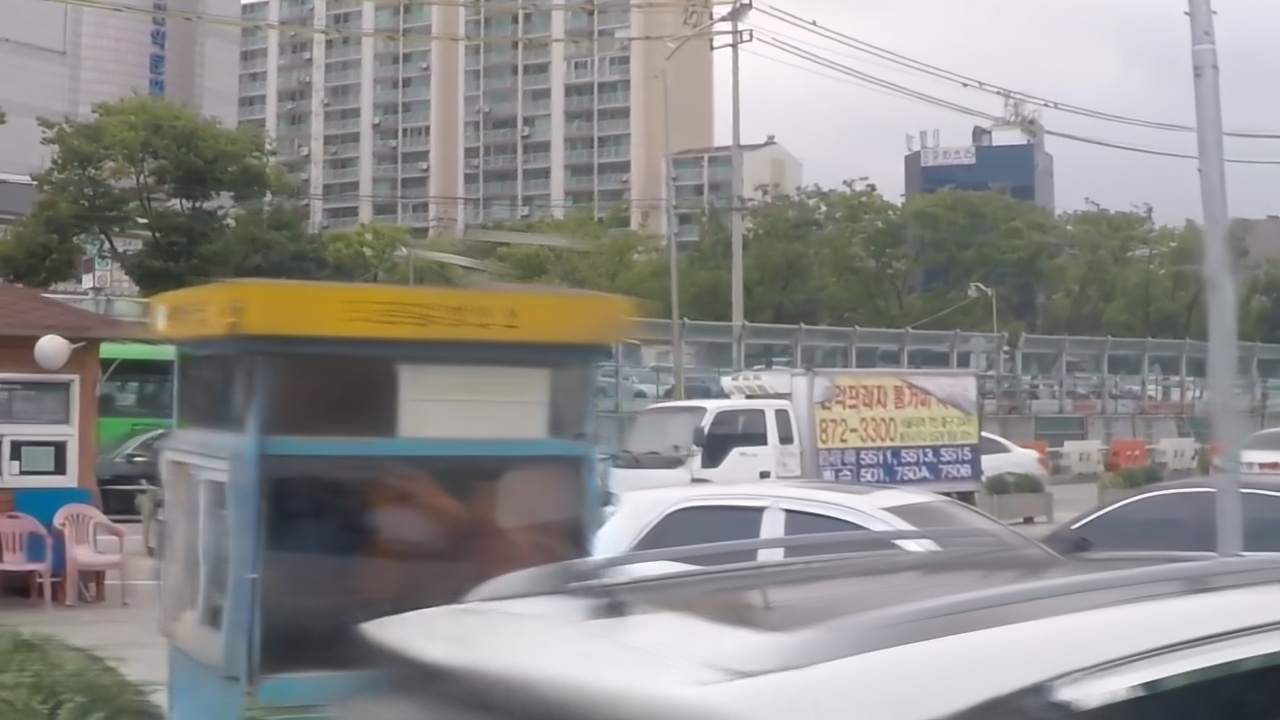}&
\includegraphics[width=0.15\textwidth, trim=200 300 700 200,clip]{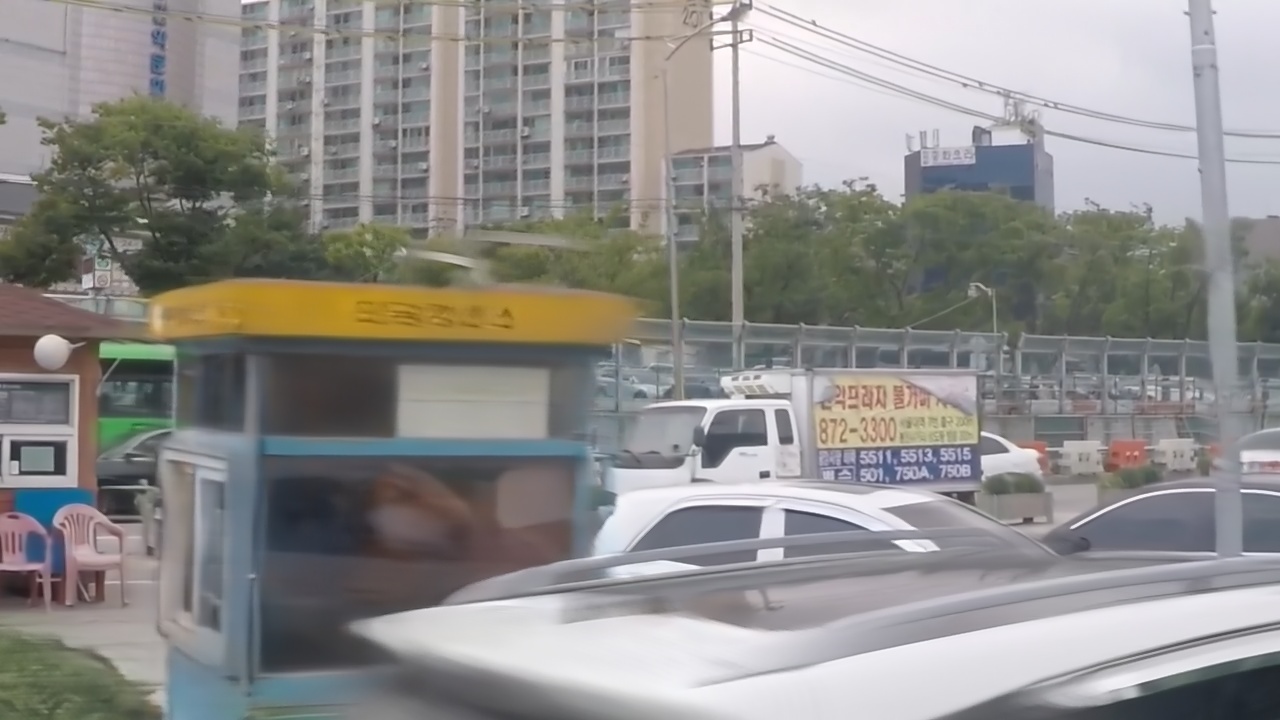}&
\includegraphics[width=0.15\textwidth,  trim=200 300 700 200,clip]{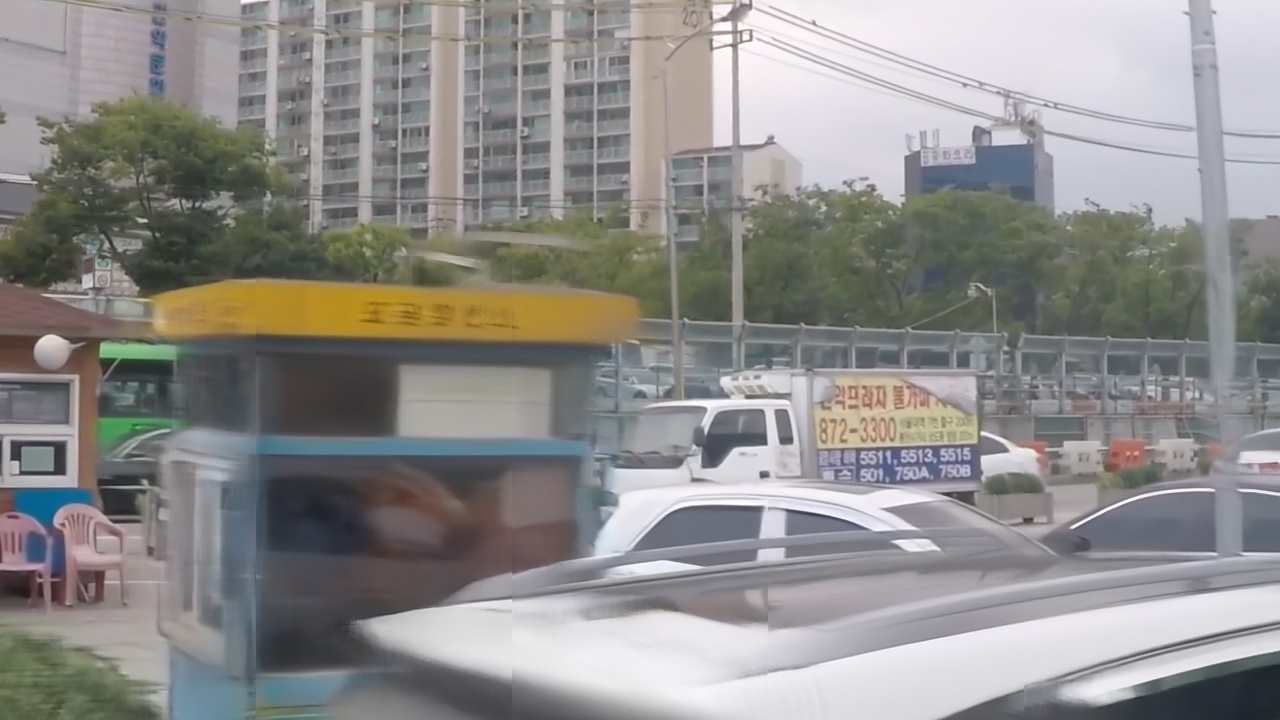}& 
\includegraphics[width=0.15\textwidth,  trim=200 300 700 200,clip]{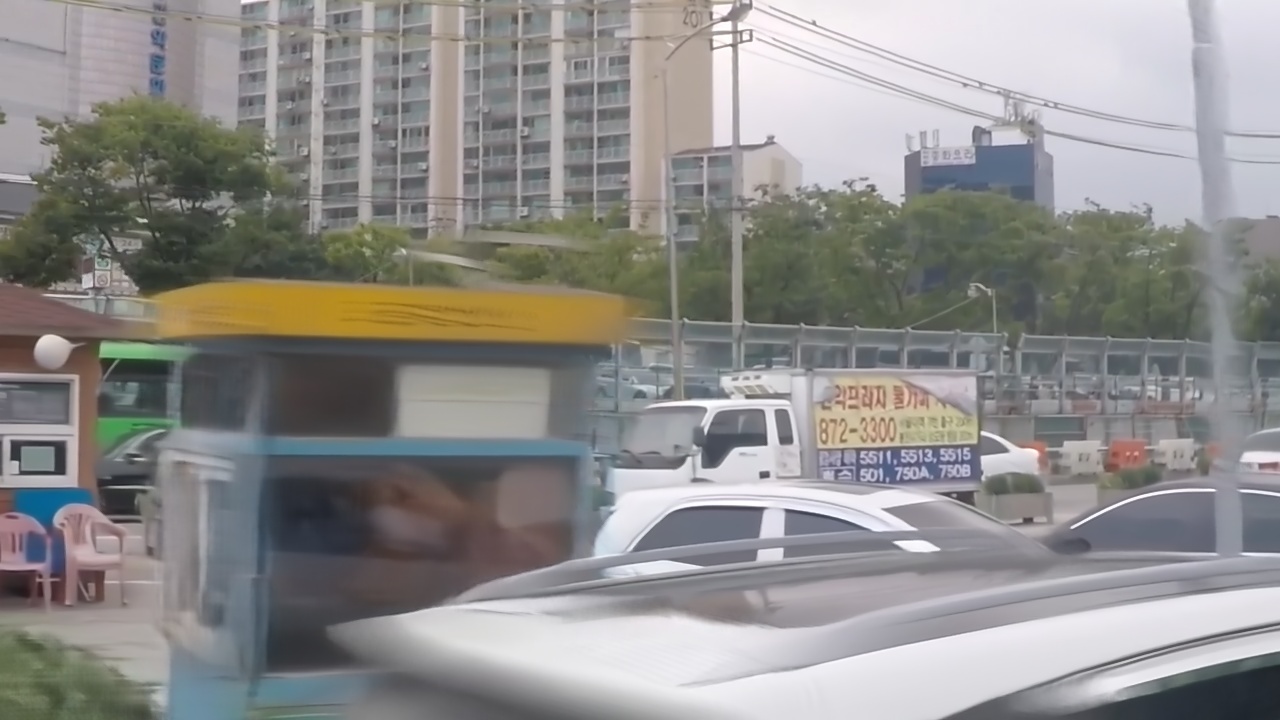}& 
\includegraphics[width=0.15\textwidth, trim=200 300 700 200,clip]{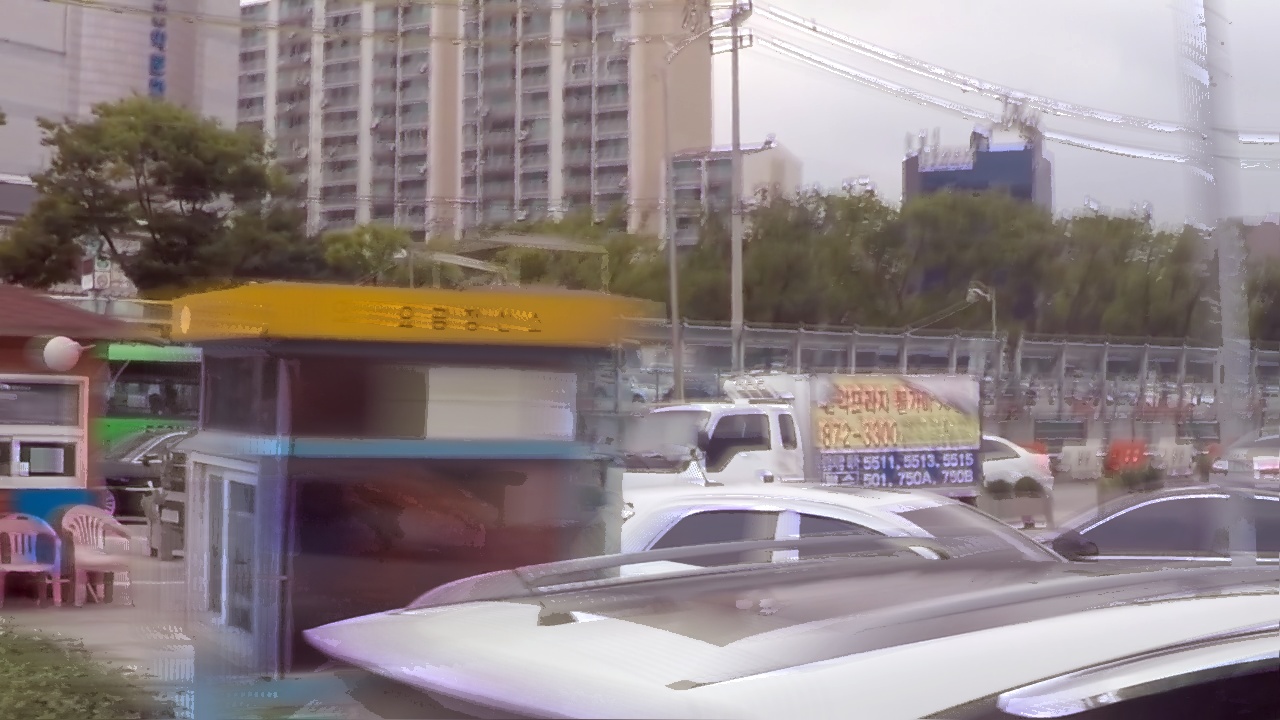}& 
\includegraphics[width=0.15\textwidth, trim=200 300 700 200,clip]{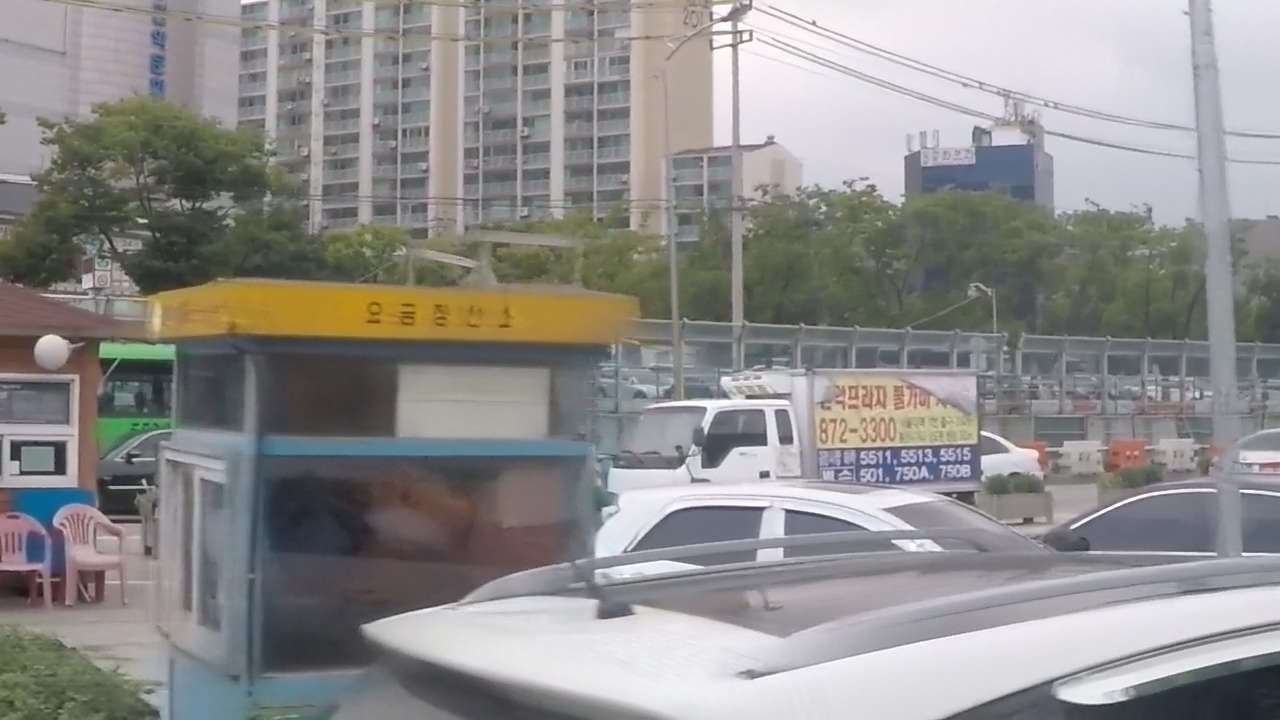}\\

\end{tabular}
\caption{\textbf{Deblurred examples on the synthetic GoPro dataset \cite{nah2017deep}.} First row: each blur image example. Second to fifth row: results obtained by \cite{huo2021blind,cho2021rethinking,chen2021hinet,zamir2021multi} image-based methods where \cite{huo2021blind} uses deformable convolutions and \cite{zamir2021multi,cho2021rethinking,chen2021hinet} use a multi-scale approach,  \cite{pan2019bringing} event-based method and our method. 
}
\label{fig:deblurringGOPRO}
\end{figure*}

\begin{figure*}[htbp]
\setlength\tabcolsep{1.5pt}
\centering

\begin{tabular}{cccccc}
\centering
  \includegraphics[width=0.14\textwidth]{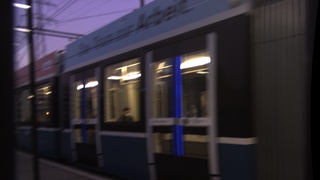} &  
  \includegraphics[width=0.14\textwidth]{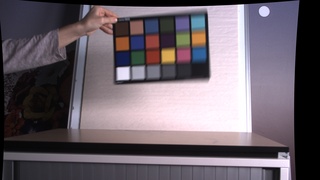} & 
  \includegraphics[width=0.14\textwidth]{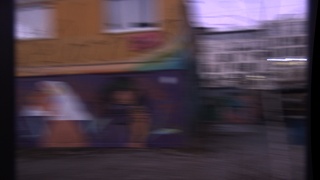} &
  \includegraphics[width=0.14\textwidth]{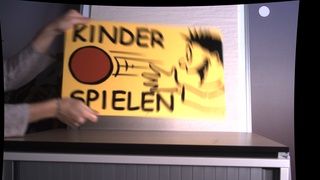} &
  \includegraphics[width=0.14\textwidth]{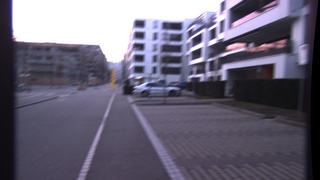}&
  \includegraphics[width=0.14\textwidth]{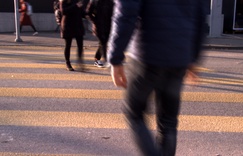}\\
\end{tabular}
\begin{tabular}{cccc}

Zamir et al. \cite{zamir2021multi} & Chen et al. \cite{chen2021hinet} & Pan et al. \cite{pan2019bringing} &  Ours \\
\includegraphics[width=0.24\textwidth, trim=25 200 400 50,clip]{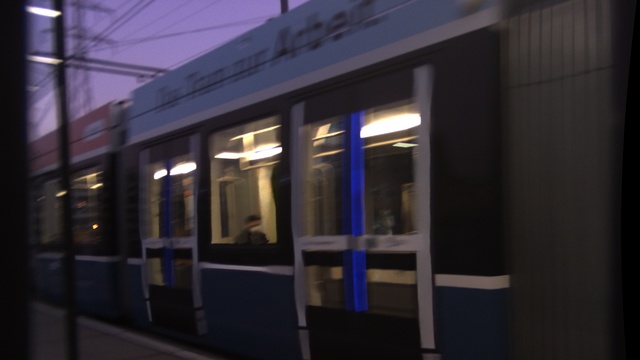}
&
\includegraphics[width=0.24\textwidth, trim=25 200 400 50,clip]{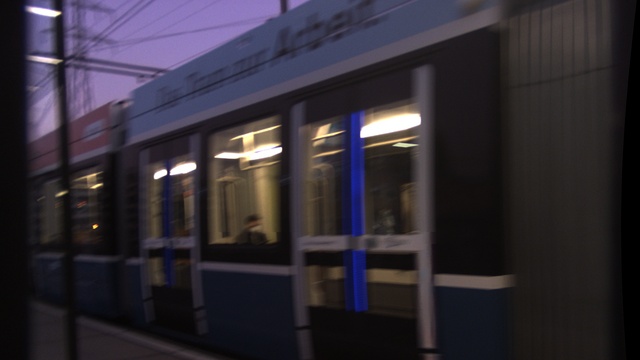}
&
\includegraphics[width=0.24\textwidth, trim=25 200 400 50,clip]{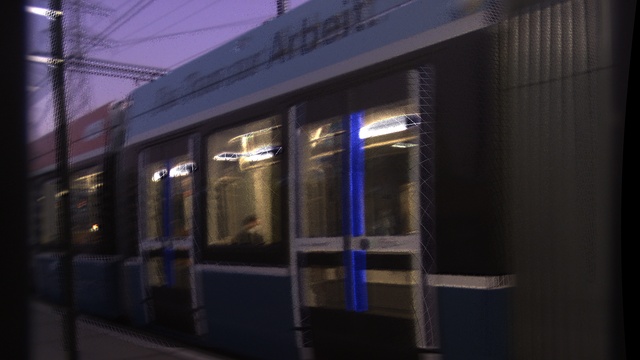}
&
\includegraphics[width=0.24\textwidth, trim=25 200 400 50,clip]{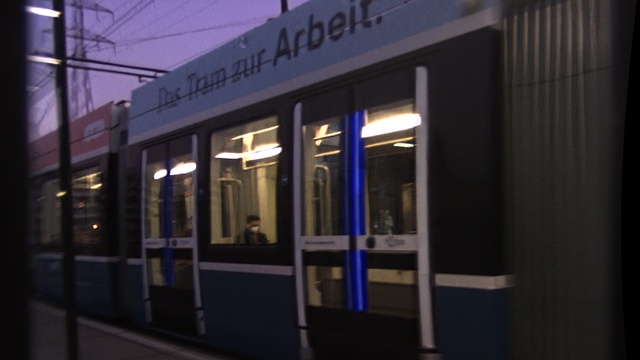}\\

\includegraphics[width=0.24\textwidth, trim=200 100 350 200,clip]{images/Results/our_dataset/low_speed_ECCV/MPRNet/10_compressed.jpeg}
&
\includegraphics[width=0.24\textwidth, trim=200 100 350 200,clip]{images/Results/our_dataset/low_speed_ECCV/HiNet/0000000024_compressed.jpeg}
&
\includegraphics[width=0.24\textwidth, trim=200 100 350 200,clip]{images/Results/our_dataset/low_speed_ECCV/Pan/0000000024_compressed.jpeg}
&
\includegraphics[width=0.24\textwidth, trim=200 100 350 200,clip]{images/Results/our_dataset/low_speed_ECCV/ours_l1/24_10_pred_compressed.jpeg}\\

\includegraphics[width=0.24\textwidth, trim=500 350 300 50,clip]{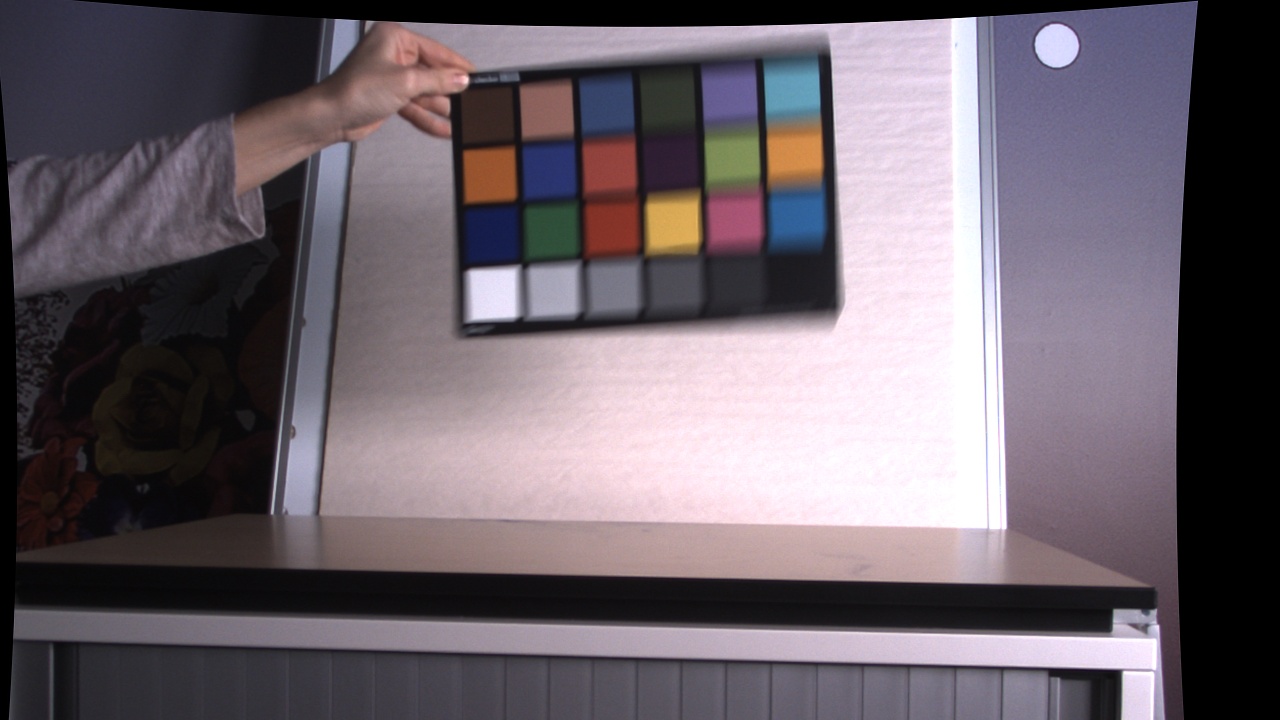}
&
\includegraphics[width=0.24\textwidth, trim=500 350 300 50,clip]{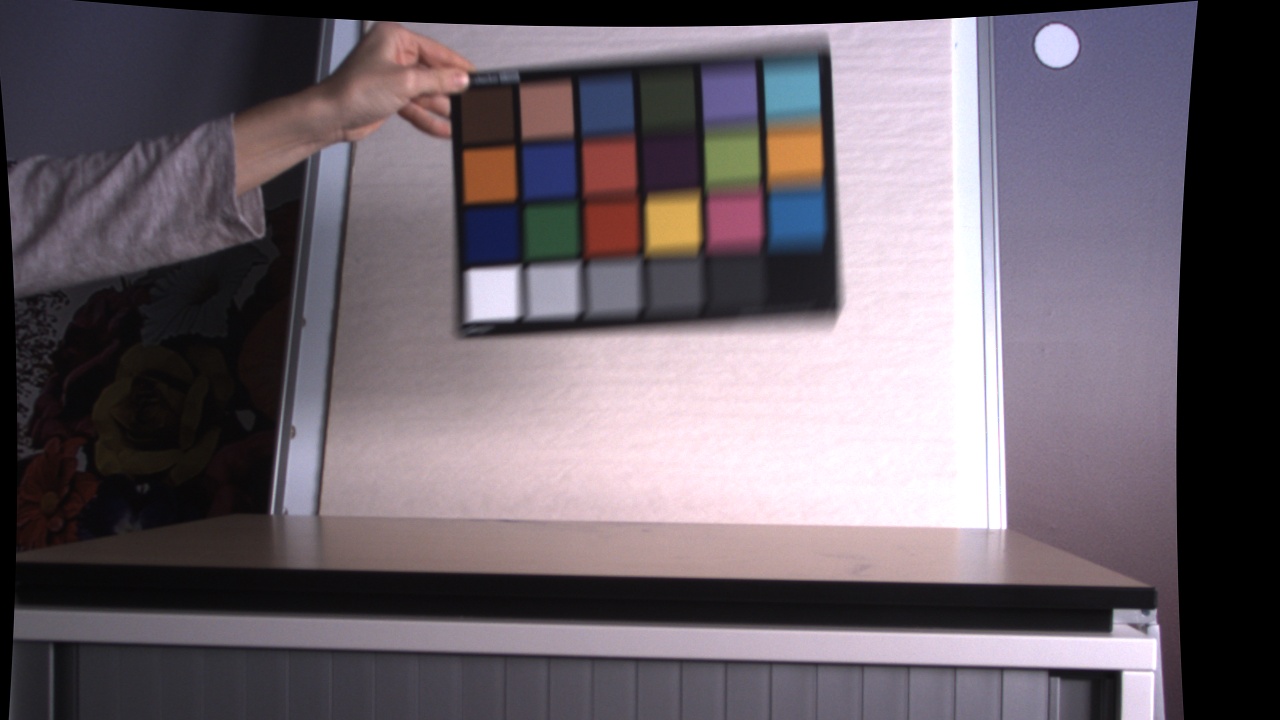}
&
\includegraphics[width=0.24\textwidth, trim=125 87 75 12,clip]{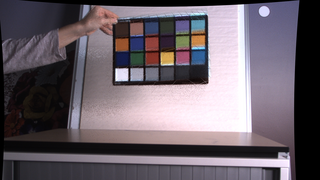}
&
\includegraphics[width=0.24\textwidth, trim=500 350 300 50,clip]{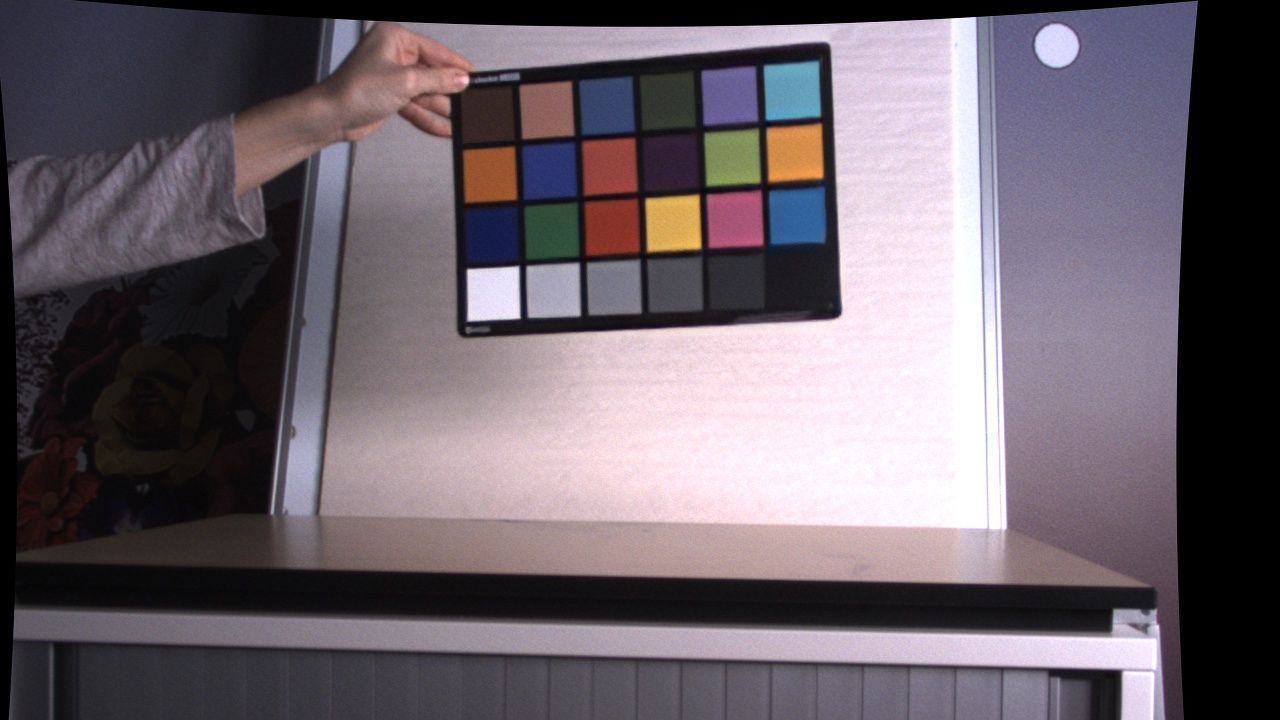}\\

\includegraphics[width=0.24\textwidth, trim=250 200 400 100,clip]{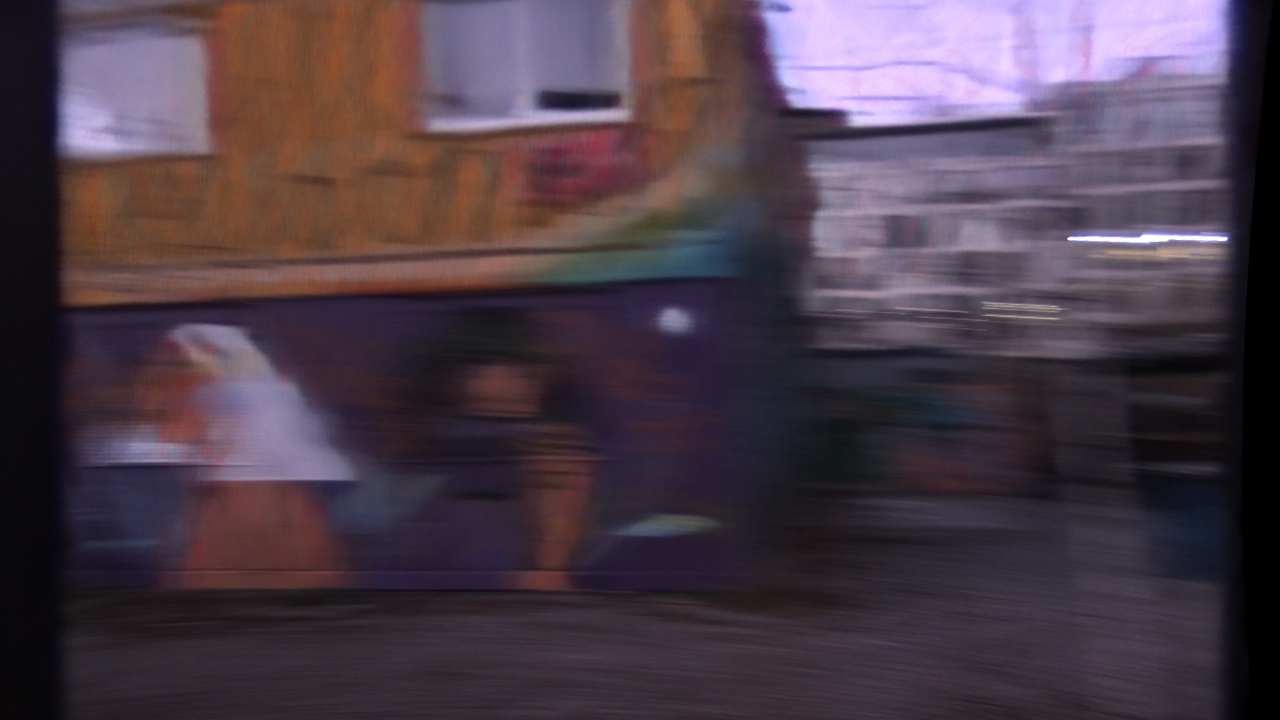}
&
\includegraphics[width=0.24\textwidth, trim=250 200 400 100,clip]{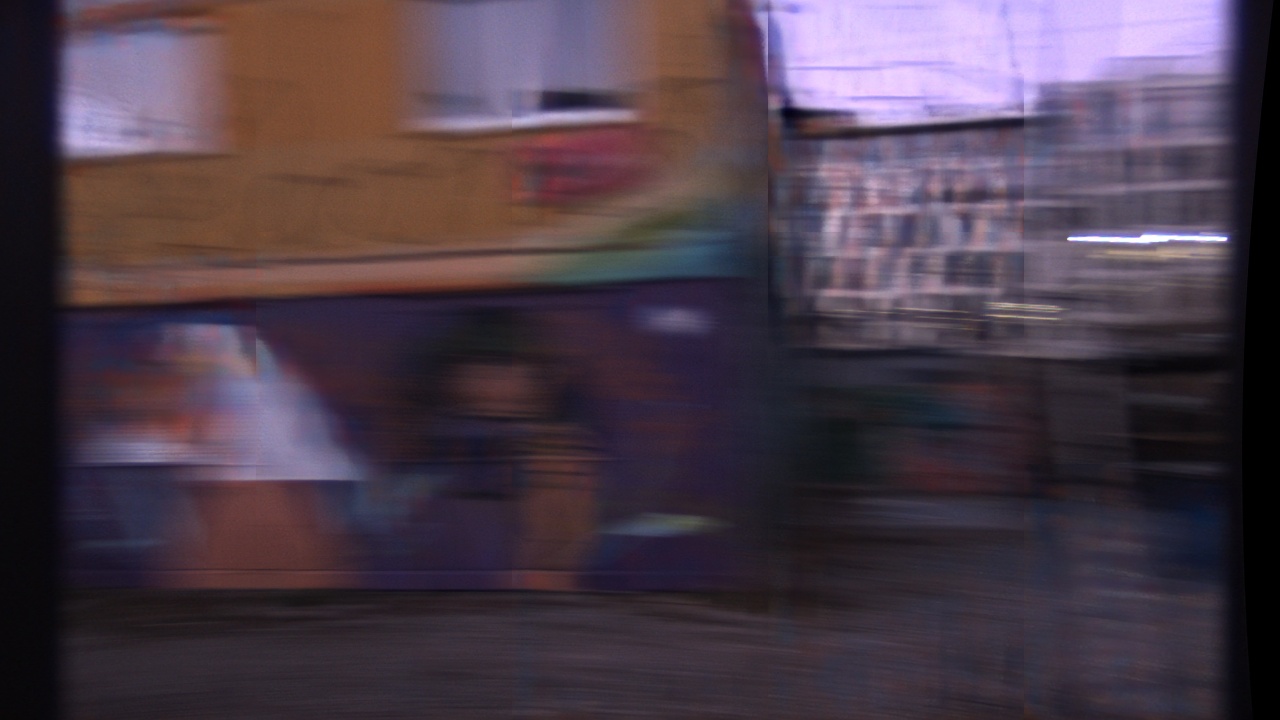}
&
\includegraphics[width=0.24\textwidth, trim=62 50 100 25,clip]{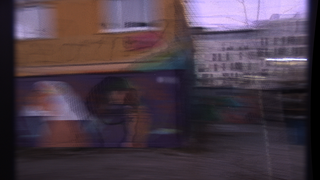}
&
\includegraphics[width=0.24\textwidth, trim=250 200 400 100,clip]{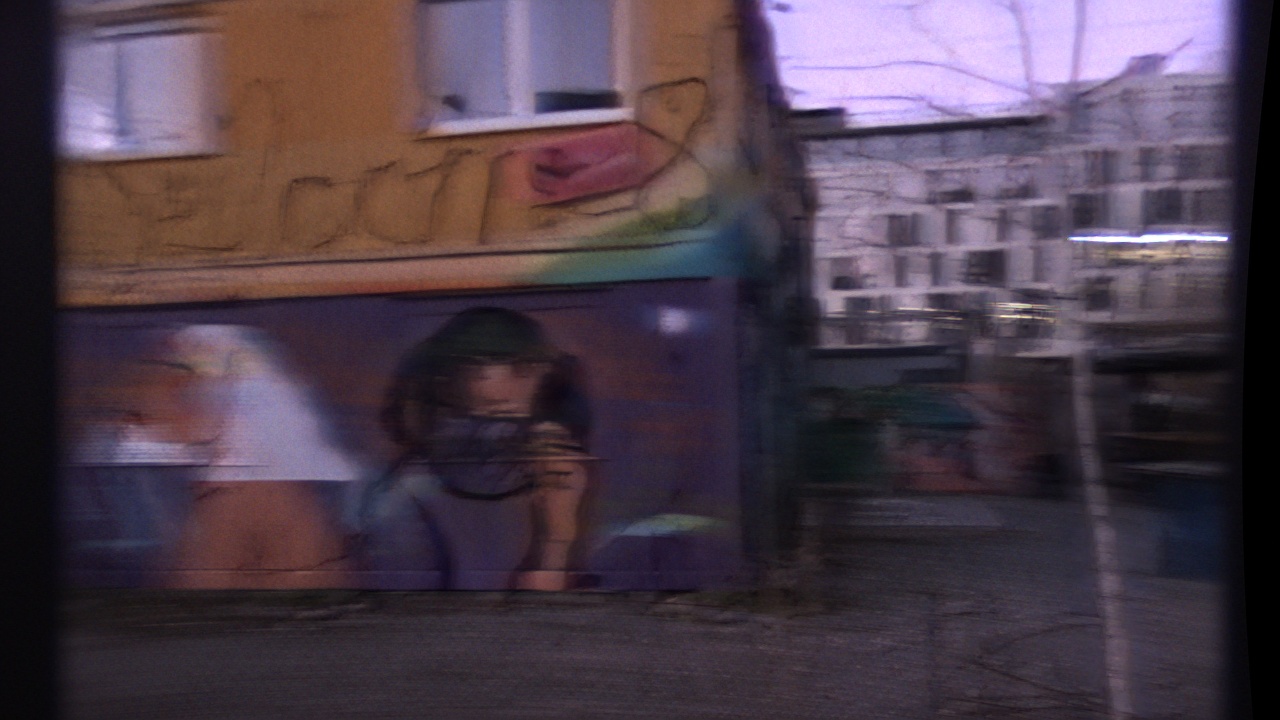} \\

\includegraphics[width=0.24\textwidth, trim=500 400 400 100,clip]{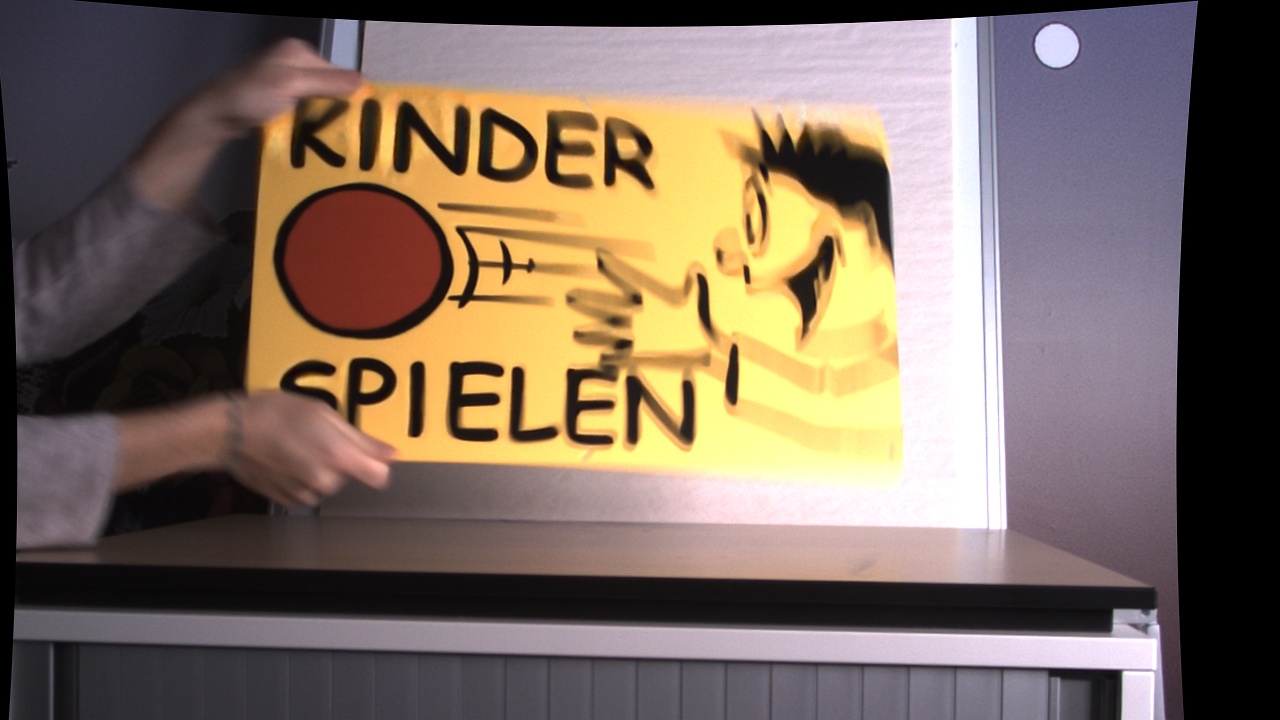}
&
\includegraphics[width=0.24\textwidth, trim=500 400 400 100,clip]{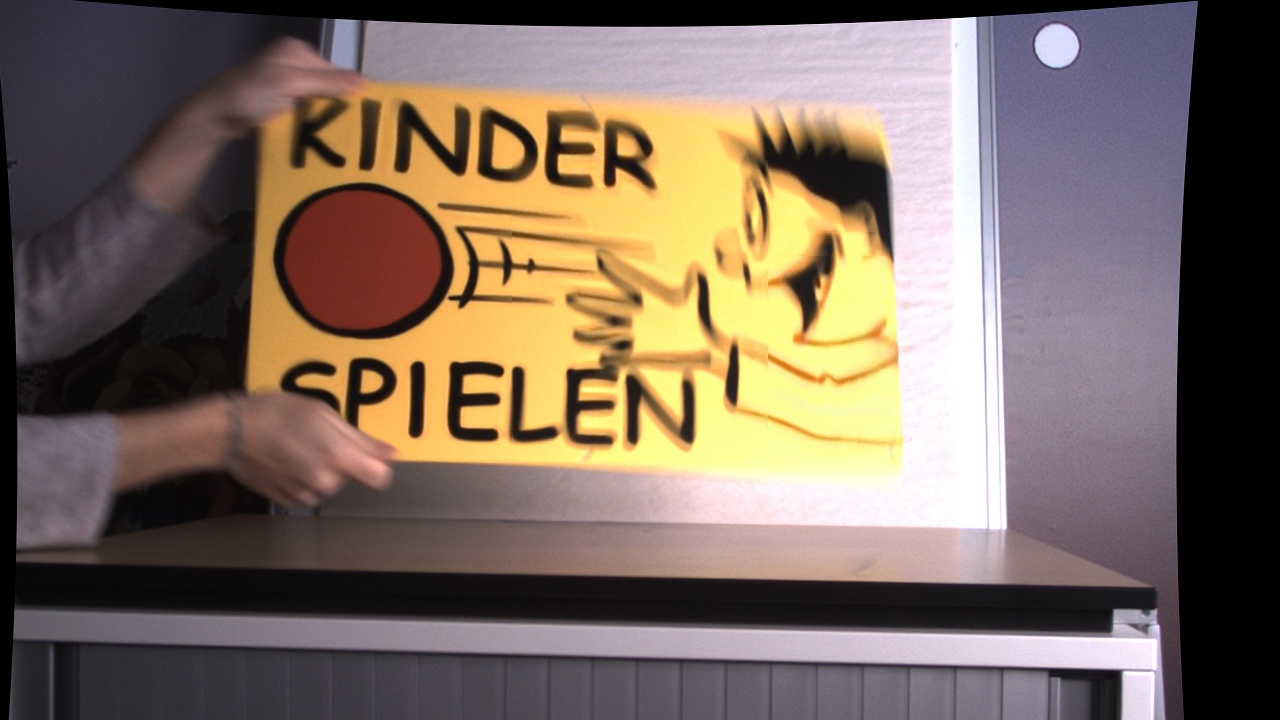}
&
\includegraphics[width=0.24\textwidth, trim=125 100 100 25,clip]{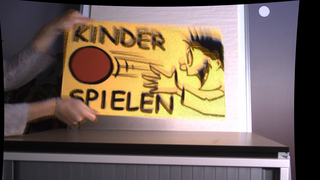}
&
\includegraphics[width=0.24\textwidth, trim=500 400 400 100,clip]{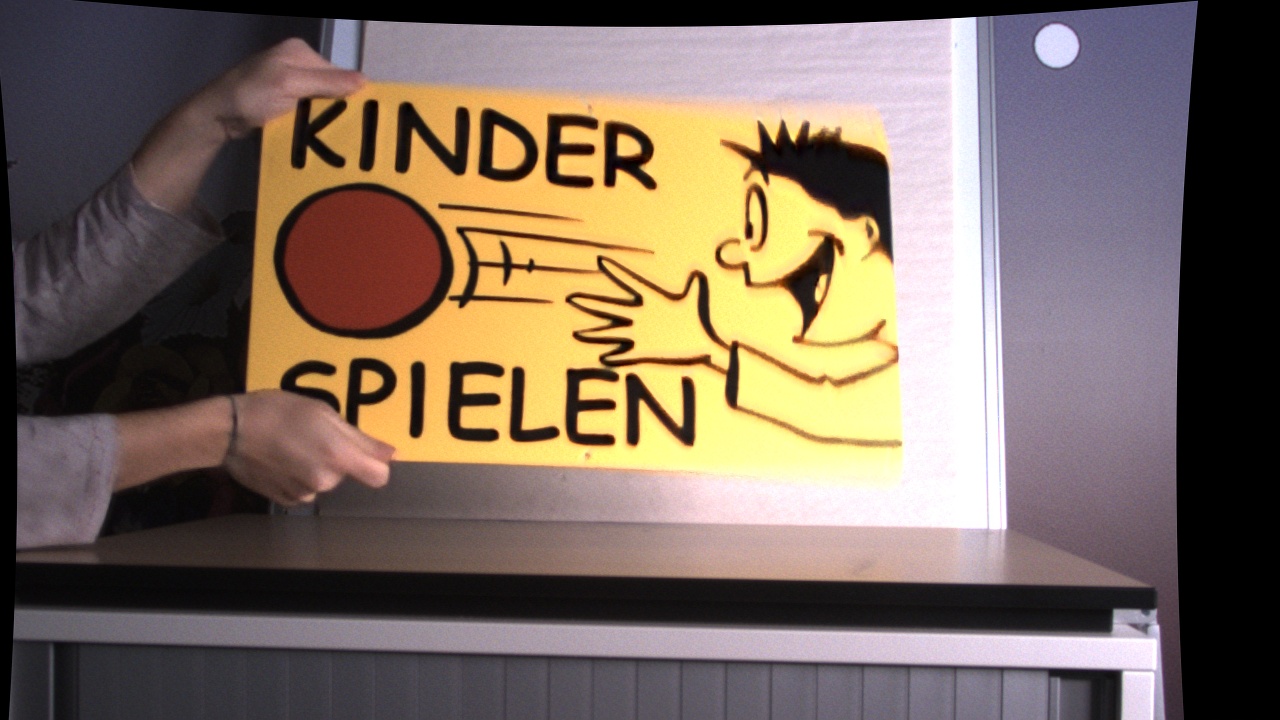}\\

\includegraphics[width=0.24\textwidth, trim=500 250 400 250,clip]{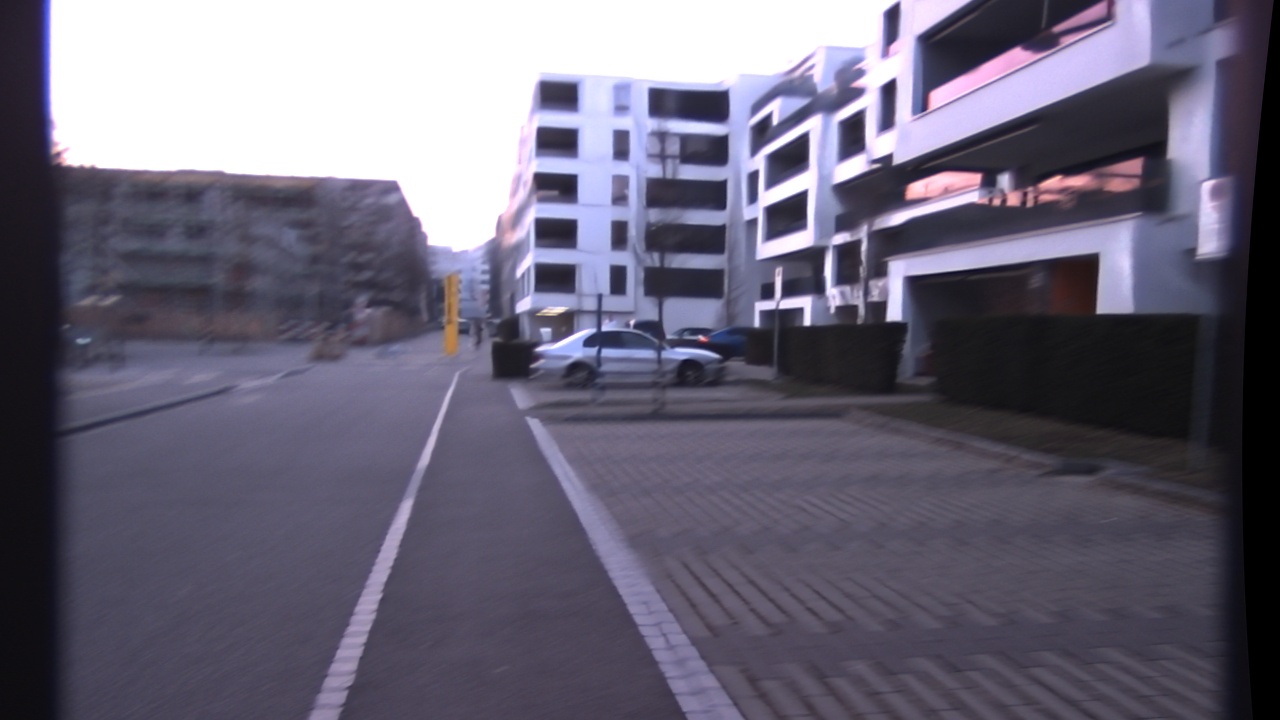}
&
\includegraphics[width=0.24\textwidth, trim=500 250 400 250,clip]{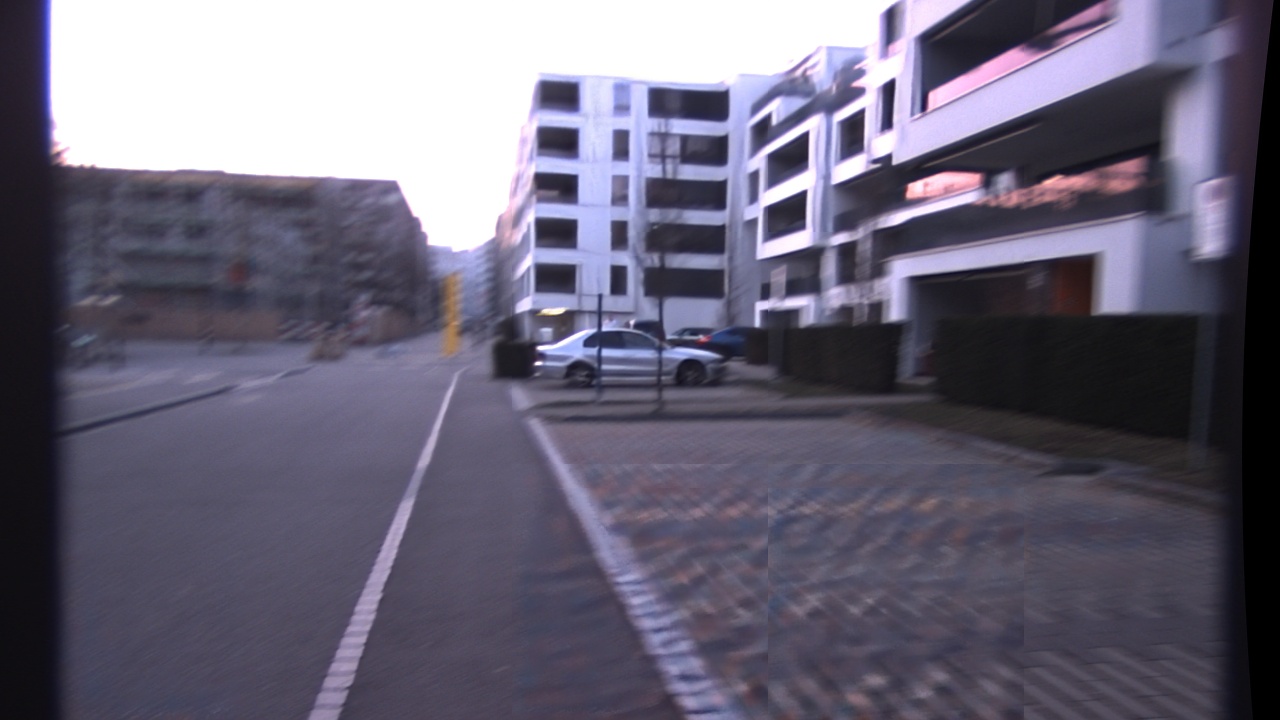}
&
\includegraphics[width=0.24\textwidth, trim=125 62 100 62,clip]{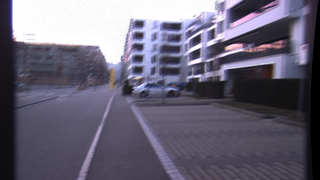}
&
\includegraphics[width=0.24\textwidth, trim=500 250 400 250,clip]{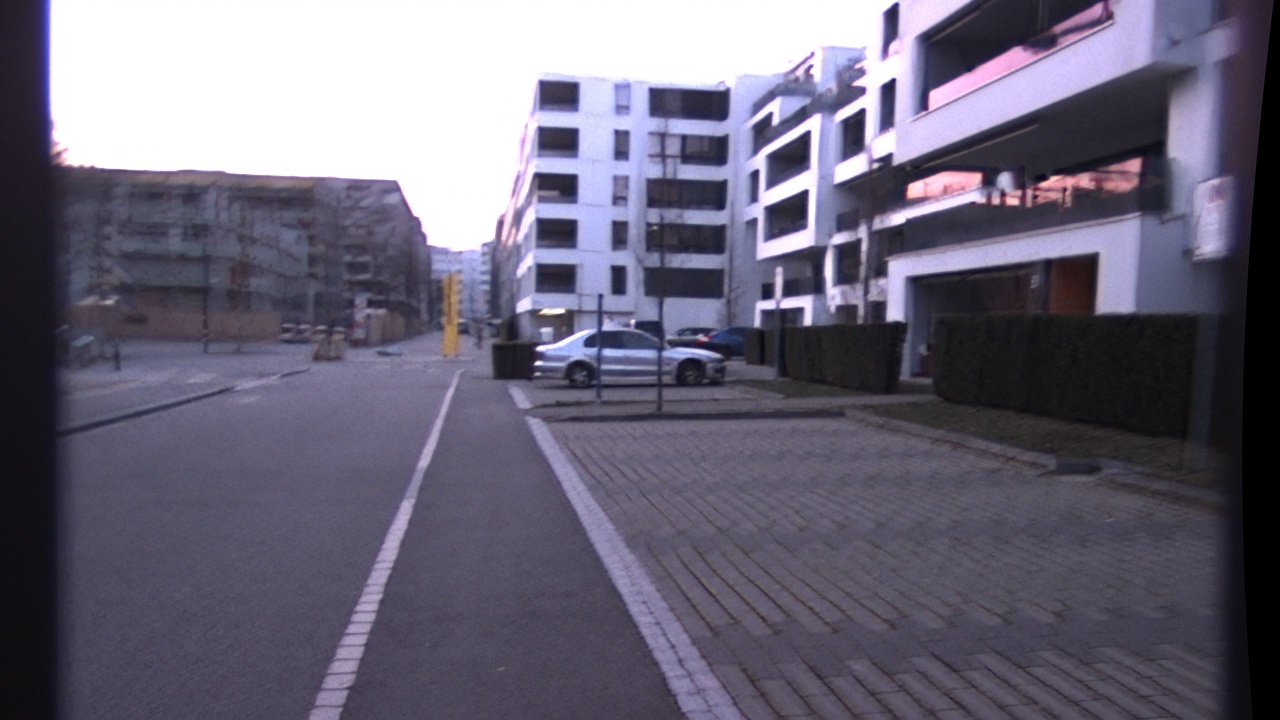}\\
	\end{tabular}
\begin{tabular}{cccc}

\includegraphics[width=0.24\textwidth, trim=100 125 175 100,clip]{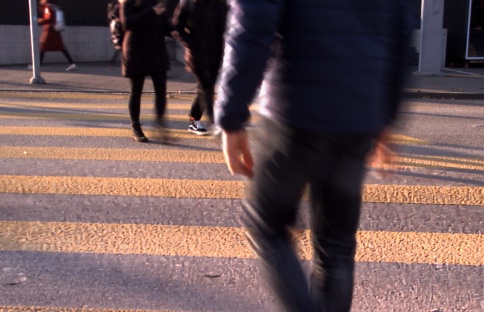}
&
\includegraphics[width=0.24\textwidth, trim=100 125 175 100,clip]{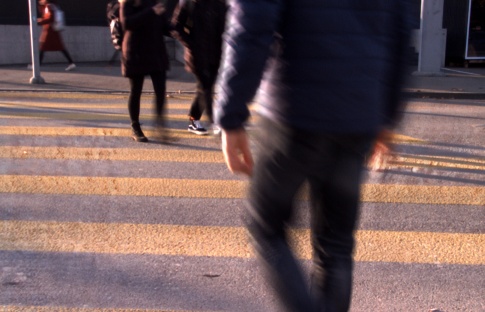}
&
\includegraphics[width=0.24\textwidth, trim=100 125 175 100,clip]{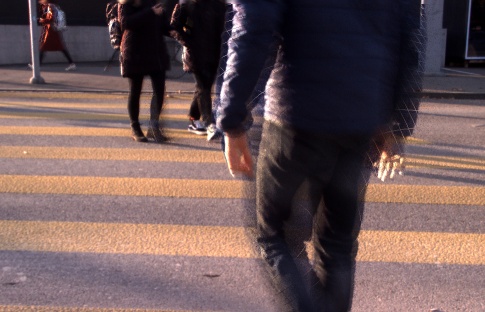}
&
\includegraphics[width=0.24\textwidth,  trim=100 125 175 100,clip]{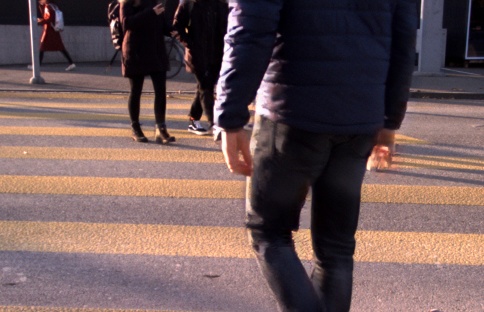}\\

	\end{tabular}
\caption{\textbf{Representative examples of state-of-the-art methods on the proposed \emph{RGBlur+E-LS}}(six first examples) and the \emph{RGBlur+E-HS} dataset (last example). First row: each blur image example. Second to eighth, from left to right: results of Zamir et al. \cite{zamir2021multi}, Chen et al. \cite{chen2021hinet}, Pan et al. \cite{pan2019bringing} and our result.}
	\label{fig:RealExamples}
\end{figure*}

\subsection{Comparison with State-of-the-Art Models}

We perform an extensive comparison with state-of-the-art \textit{image} deblurring methods.
We re-ran experiments if code is available for the best-performing methods. Our comparison includes model-based \cite{xu2013unnatural,hyun2013dynamic,whyte2012non,gong2017motion}, learning-based \cite{jin2018learning,kupyn2018deblurgan,nah2017deep,zhang2018dynamic,kupyn2019deblurgan,yuan2020efficient,zhang2019deep,tao2018scale,gao2019dynamic,park2020multi,suin2020spatially,huo2021blind,cho2021rethinking,zamir2021multi,chen2021hinet} and event-based methods \cite{pan2019bringing,pan2020high,jiang2020learning,haoyu2020learning,zhang2020hybrid}. The comparison is twofold: first, as widely adopted in the literature, we compare quantitatively and qualitatively with the GoPro synthetic dataset. Second, we compare quantitatively with the first subset of the RGBlur+E dataset and qualitatively with the second subset. 

\textbf{Comparison on the GoPro dataset.}
Tab.~\ref{tab:resultsGoPro} reports the results on the synthetic GoPro dataset. Gray values are directly extracted from papers. Our approach achieves the best performance in terms of both PSNR and SSIM improving the PSNR by $+5,94$,  $+1,57$, and $+2,08$ dB compared with model-, learning-, and event-based methods, respectively.
We show in Fig.~\ref{fig:deblurringGOPRO} a visual comparison between our method and several state-of-the-art methods.
The proposed method recovers the sharpest details in various challenging situations, \emph{e.g.} details on the faces, like eyes, nose, and eyebrows or the car plate numbers. Also, our method can handle low-contrast situations where no events are present, \emph{e.g.} the fingers in the moving hand in the first example, or severe blur situations.

\textbf{Comparison on the proposed RGBlur+E Dataset}
One of the main drawbacks of state-of-the-art methods is the lack of generalization to real blur and events. In this section, we evaluate the performance in real case scenarios. 
We use RGBlur+E-HS to quantitatively evaluate the performance of event-based methods to real event data. Note that, this set contains real events but synthetically generated blurry images by averaging $11$ frames.
To fairly compare image-based method where the entire input is all synthetic, we finetune the our model with the training subset of RGBlur+E-HS.
We report quantitative results in Tab.~\ref{tab:resultsRGBlur}. Our model outperform all other methods while encountering real events.
Examples of this set can be found in the last row in Fig.~\ref{fig:RealExamples} and in the Supp. Mat.. 

\begin{table}[h!bt]
    \centering
    \begin{tabular}{l|c|c|c|c|c}
\hline
\textbf{Method} & Zamir et al. \cite{zamir2021multi}  & Chen et al. \cite{chen2021hinet} & Pan et al. \cite{pan2019bringing} &  Ours  & Ours Finetuned \\ 
\hline
\textbf{PSNR}  $\uparrow$  & 33.16 & 32.67 & 24.20 & 31.95 & \textbf{34.24} \\
\textbf{SSIM}  $\uparrow$  & 0.86  & 0.86 & 0.76 & 0.85 & \textbf{0.87}\\
\hline
    \end{tabular}
    \caption{\textbf{RGBlur+E-HS performance.} }
    \label{tab:resultsRGBlur}
\end{table}

\section{Conclusions}
In this work, we propose a novel deep learning architecture that decouples the estimation of the per-pixel motion from the deblur operator. First, we dynamically learn offsets and masks containing motion information from events. Those are used by modulated deformable convolutions that will deblur image and event features at different scales.
Moreover, we reconstruct the sharp image in a coarse-to-fine multi-scale approach to overcome the inherent sparsity and noisy nature of events.
Additionally, we introduce the \emph{RGBlur+E} dataset, the first dataset containing real HR RGB images plus events.
Extensive evaluations show that the proposed approach achieves superior performance compared to state-of-the-art image and event-based methods on both synthetic and real-world datasets. 

\clearpage
\bibliographystyle{splncs04}
\bibliography{main}
\end{document}